\documentclass{article}

\PassOptionsToPackage{numbers, compress}{natbib}

\usepackage[preprint]{neurips_2024}
\usepackage{times}
\usepackage{soul}
\usepackage{url}
\usepackage[hidelinks]{hyperref}
\usepackage[utf8]{inputenc}
\usepackage[small]{caption}
\usepackage{graphicx}
\usepackage{amsmath}
\usepackage{amsthm}
\usepackage{booktabs}
\usepackage{algorithm}
\usepackage{algorithmic}
\usepackage{graphicx} 
\usepackage{xcolor}
\usepackage{etoc}
\usepackage{wrapfig}
\usepackage{amssymb}
\usepackage{amsmath}
\usepackage{tabularx}
\usepackage{adjustbox}
\usepackage{bbm}
\usepackage{wrapfig}
\usepackage{multicol}
\usepackage{multirow}

\usepackage{listings}
\usepackage{color}

\definecolor{dkgreen}{rgb}{0,0.6,0}
\definecolor{gray}{rgb}{0.5,0.5,0.5}
\definecolor{mauve}{rgb}{0.58,0,0.82}

\definecolor{darkgoldenrod}{rgb}{0.72, 0.53, 0.04}
\definecolor{indianred}{rgb}{0.8, 0.36, 0.36}
\definecolor{mediumseagreen}{rgb}{0.24, 0.7, 0.44}
\definecolor{mediumpurple}{rgb}{0.58, 0.44, 0.86}

\definecolor{ao(english)}{rgb}{0.0, 0.5, 0.0}

\usepackage[utf8]{inputenc} 
\usepackage[T1]{fontenc}    
\usepackage{hyperref}       
\usepackage{url}            
\usepackage{booktabs}       
\usepackage{amsfonts}       
\usepackage{nicefrac}       
\usepackage{microtype}      
\usepackage{xcolor}         

\newcommand\blfootnote[1]{%
  \begingroup
  \renewcommand\thefootnote{}\footnote{#1}%
  \addtocounter{footnote}{-1}%
  \endgroup
}

\title{Code Simulation Challenges for Large Language Models}

\author{%
  Emanuele La Malfa\thanks{First author. Correspondence at emanuele.lamalfa@cs.ox.ac.uk.}$  \ \ ^{\dagger}$ \\
  Department of Computer Science\\
  University of Oxford\\
  \And
  Christoph Weinhuber \\
  Department of Computer Science\\
  Technical University of Munich \\
  \And
  Orazio Torre \\
  University of Salerno \\
  \And
  Fangru Lin \\
  Faculty of Linguistics, Philology, and Phonetics\\
  University of Oxford \\
  \And
  Samuele Marro \\
  Department of Computer Science and Engineering\\
  University of Bologna \\
  \AND
  Anthony Cohn${^\dagger}$ \\
  Faculty of Engineering and Physical Sciences\\
  University of Leeds \\
  \And
  Nigel Shadbolt${^\dagger}$ \\
  Department of Computer Science\\
  University of Oxford \\
  \And
  Michael Wooldridge${^\dagger}$ \\
  Department of Computer Science\\
  University of Oxford\\
}

\begin{document}

\maketitle

\begin{abstract}
Many reasoning, planning, and problem-solving tasks share an intrinsic algorithmic nature: correctly simulating each step is a sufficient condition to solve them correctly.\blfootnote{\noindent$^\dagger$The author is also affiliated and supported by The Alan Turing Institute, London.}
This work studies to what extent Large Language Models (LLMs) can simulate coding and algorithmic tasks to provide insights into general capabilities in such algorithmic reasoning tasks. We introduce benchmarks for straight-line programs, code that contains critical paths, and approximate and redundant instructions. We further assess the simulation capabilities of LLMs with sorting algorithms and nested loops and show that a routine's computational complexity directly affects an LLM's ability to simulate its execution. 
While the most powerful LLMs exhibit relatively strong simulation capabilities, the process is fragile, seems to rely heavily on pattern recognition, and is affected by memorisation. We propose a novel off-the-shelf prompting method, Chain of Simulation (CoSm), which instructs LLMs to simulate code execution line by line/follow the computation pattern of compilers. CoSm efficiently helps LLMs reduce memorisation and shallow pattern recognition while improving simulation performance. We consider the success of CoSm in code simulation to be inspirational for other general routine simulation reasoning tasks.\footnote{The code is available here: \url{https://github.com/EmanueleLM/CodeSimulation}}

\end{abstract}

\section{Introduction}
\begin{figure}
\centering
\includegraphics[width=0.9\linewidth]{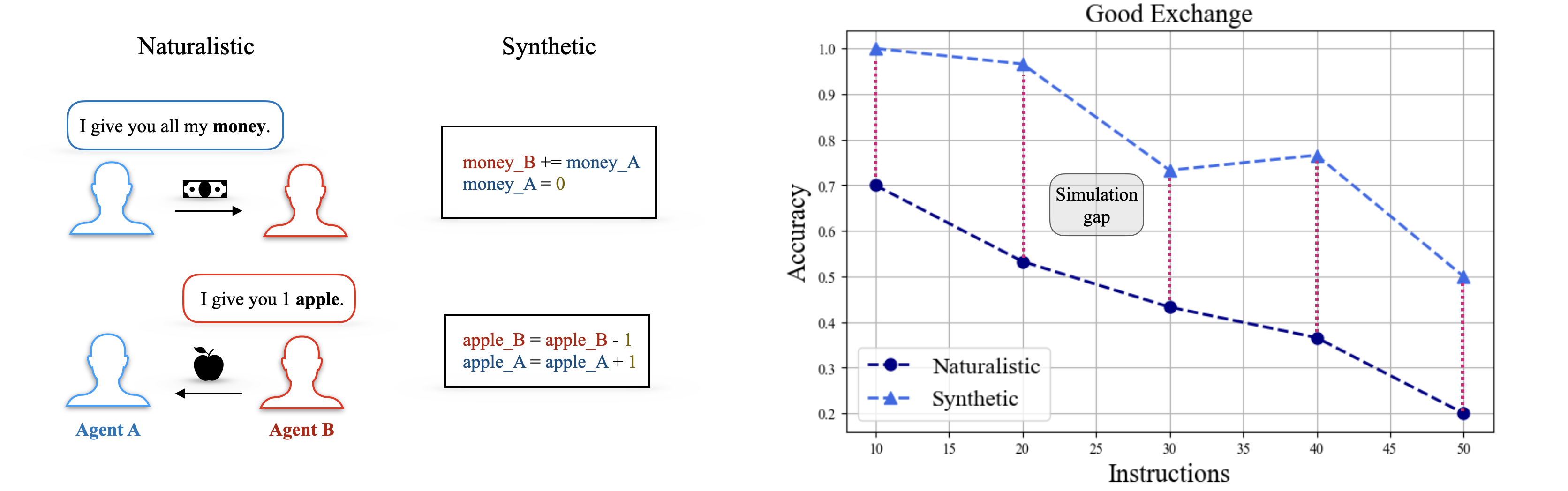}
\caption{Left: an example of the \emph{naturalistic} vs. \emph{synthetic} good exchange settings. The former describes, in natural language, two agents who exchange goods; the latter is an equivalent formulation in code. GPT-3.5-Turbo performances on the tasks are correlated (right), but it performs better on the \emph{synthetic} task (a ``simulation gap''). We conduct experiments on $30$ samples per instruction class with $\{10, 20, 30, 40, 50\}$ interactions/ lines of code. 
}
\label{fig:motivating-example}
\vspace{-3mm}
\end{figure}

A major area of interest at the time of writing is to understand the capabilities of Large Language Models (LLMs) beyond the tasks of natural language understanding and generation for which they were originally designed. 
Many benchmarking tasks, including Theory of Mind~\cite{pmlr-v80-rabinowitz18a}, planning~\cite{hao2023reasoning,ouyang2022training}, and high-order reasoning~\cite{webb2023emergent}, necessitate turning a prompt, expressed in natural language, into a procedure that must then be carried on faultlessly.  
Consider a problem where two agents interact and exchange goods, as depicted in Figure~\ref{fig:motivating-example}. An LLM prompted to compute the number of goods at the end of an iteration should be able to sum and assign variables correctly. Such a \emph{naturalistic} problem has an intrinsic algorithmic nature as \emph{synthetic} code (Figure~\ref{fig:motivating-example}, centre).
Our preliminary experiments, which we will revise in Section~\ref{sec:cosm} and pair others in literature~\cite{lin2024graphenhanced}, evidence a strong performance correlation between \emph{naturalistic} and \emph{synthetic} tasks. 

Beyond the example above, the code simulation capabilities of LLMs are an important object of study for at least two complementary reasons. First, LLMs have shown some capabilities for planning~\cite{ouyang2022training}, which requires reasoning step-by-step and recursively dividing a problem into its elementary components. However, such compositional reasoning performance is highly fragile with trivial prompt variations~\cite{turpin2024language}, so it is unclear how competent LLMs are in algorithmic reasoning. With highly structured input, code simulation can provide an intuition about this.
Second, code simulation requires an LLM to turn instructions formulated in code and natural language into a procedure or an optimisation problem that the model must solve correctly, and thus answers the question of whether LLMs can serve as \emph{digital} computational models~\cite{jojic2023gpt}.

In more detail, in this paper:
\begin{itemize}
    \item We leverage the analogy between LLMs and CPUs to study the former as \emph{analog} simulators of \emph{digital} devices on algorithms. We test the ability of LLMs on \textbf{six coding benchmarks} (ref. Methodology), on GPT-3.5-Turbo, GPT-4~\cite{OpenAI2023GPT4TR}, Jurassic2-Ultra~\cite{lieber2021jurassic}, LLaMA3-70B~\cite{llama3modelcard}, LLaMA2-70B~\cite{touvron2023llama2} and CodeLLaMA-34b-Instruct~\cite{roziere2023code}, with standard promoting techniques including Chain of Thought~\cite{wei2022chain}, Self-consistency~\cite{wang2023selfconsistency} and k-shot prompting~\cite{kojima2022large}. We identify GPT-4 and LLaMA3 as the most powerful models, with the former being better at leveraging pattern matching and the latter at straight code simulation.
    \item We propose an off-the-shelf prompting technique, Chain of Simulation (CoSm), that improves performance on many tasks and suppresses memorisation when it is detrimental to the downstream task. CoSm \textbf{elicits simulation} of each instruction sequentially while outputting the program trace and is seamlessly integrated with techniques such as Chain of Thought and Self-consistency.
\end{itemize}

In summary, our work contributes to benchmarking \emph{bare-bones} LLMs~\cite{shanahan2022talking} 
, i.e., those not explicitly integrated with compilers or code interpreters, and shows how the most capable models exhibit simulation capabilities. Our work makes falsifiable the hypothesis that LLMs can simulate basic instructions with low error probability analogously to digital computers~\cite{jojic2023gpt}, and provides empirical evidence that LLMs are not even ``Chinese rooms''~\cite{searle1980minds}, a widely discussed argument in philosophy of mind.

\section{Related Work} 
\paragraph{Code generation and compositionality.}
LLMs to understand, generate, and improve code have been mainly developed to produce debugging information without invoking a compiler/interpreter~\cite{hou2023large,santos2023always,chen2021evaluating,widjojo2023addressing,zan2023large}.
Code generation and simulation require some degree of compositionality~\cite{mccoy2023much}, i.e., the result of complex expressions can be determined by their constituents and the rules used to combine them.
Recent works explored compositionality in terms of simple mathematical operations that LLMs can execute~\cite{frieder2023mathematical,yang2023code,yuan2023well}, and revealed how the most potent models do not achieve that~\cite{mccoy2023embers,west2023generative}.
Our work further explores the tension between memorisation and performance on complex tasks~\cite{berglund2023reversal,eldan2023s,yang2023code}, with results that illustrate how the former is at tension with the size of a model, the so-called ``inverse scaling law''~\cite{biderman2023emergent}.

\paragraph{Code simulation.}
Before the breakthrough of closed-source LLMs~\cite{la2023arrt}, a seminal work tested LLMs on code simulation, showing how keeping track of the variables improves their capabilities~\cite{nye2021show}. Successive works explored LLMs and code simulation~\cite{chen2024language,tufano2023predicting,zhou2023algorithms}, particularly in~\cite{liu2023code}, where the authors fine-tune Transformer-based models to output the program trace of a code snippet. 
Recent developments in this field go under the name of ``code reasoning'', as a model's ability to predict a variable's state at runtime~\cite{chen2024evaluating}, the output of a statement/function~\cite{gu2024cruxeval,liu2024codemind}, or their capability to handle recursion~\cite{zhang2024transformerbased}.
At the architectural level, several works studied Transformers and attention-based models regarding the operations and programming languages they interpret and execute~\cite{weiss2021thinking} and their recursive code simulation capabilities~\cite{zhang2023can}.
On a broader perspective, past work investigated the Turing-completeness of LLMs~\cite{giannou2023looped,perez2021attention,schuurmans2023memory,wei2022statistically}, and their ability to follow instructions~\cite{ouyang2022training} and policies expressed as code~\cite{liang2023code}.

\section{Methodology} 
Formally, a Language Model is defined as a function that predicts the next token (out of a finite vocabulary) conditioned on the sequence of previously fed/generated tokens, namely $\psi: V^* \xrightarrow{} \mathbb{P}(V)$. 
In our setting, a problem is specified as a tuple $(x, p)$, where $x$ instructs the model to simulate $p$, a program coded in a target language (e.g., Python, pseudo-code, etc.). The output of the LLM is then compared for correctness against that of a compiler/interpreter $\Omega$ for $p$, i.e., $\Omega(p)$.
We select Python3 as our programming language for $p$ as it represents the language of reference in most LLMs such as Code-LLaMA~\cite{roziere2023code} and is among the most covered languages in open-source LLMs~\cite{gao2020pile,scao2022bloom} and network Q\&A platforms such as Stack Exchange. Furthermore, its syntax, for simple programs, resembles that of pseudo-code, thus abstracting from complex programming constructs. 

Metrics such as the accuracy, expressed as $\frac{1}{N}\sum_{i=1}^{N}\mathbbm{1} [\psi(y_i | x,p_i) = \Omega(p_i)]$, inform us as to the simulation capabilities of a model.
Another metric we consider is how \emph{wrong} a model is in its prediction, i.e., the average distance between the correct result and the model output, namely $\frac{1}{N}\sum_{i=1}^{N} |\psi(y_i | x,p_i) - \Omega(p_i)|$.
On tasks that require simulating multiple independent instructions or returning multiple correct predictions, we measure the prediction error as the Levenshtein distance between the prediction and the ground truth (as tuples), namely $\frac{1}{N}\sum_{i=1}^{N} |\psi(y_i | x,p_i) \cap \Omega(p_i)|$.\footnote{We assume w.l.o.g. that $y$ contains the relevant output of the program and can be directly compared to $\Omega(p)$.}
We also analysed the LLMs' responses to identify the most common reasons for failure in code simulation.

In summary, our methodology addresses whether LLMs can reliably simulate code execution, from basic operations such as addition and assignment to more complex functions such as nested loops and sorting algorithms. 
To achieve this, we introduce six benchmarks: (i) \textbf{straight-line code simulation}, which tests the ability of a program to simulate simple, sequential instructions consistently; (ii) \textbf{smart execution}, i.e., straight-line programs whose portion of the program relevant to output the correct answer is encapsulated in a shorter, straight-line program; (iii) \textbf{approximate computation}, which tests the capacity of an LLM to perform multiple independent sub-programs; (iv) \textbf{redundant algorithms}, where a model is fed multiple equivalent programs and informed that such programs are expected to return the same result, (v) \textbf{nested loops}, to connect computational complexity with a model's simulation capabilities, and (vi) \textbf{sorting algorithms} of varying complexity, both in the recursive and iterative versions.

We conducted our experiments on GPT-3.5-Turbo, GPT-4, Jurassic2-Ultra, LLaMA2-70B, LLaMA3-70B and CodeLLaMA-34b-Instruct (we report the parameters used to invoke each LLM in Appendix~\ref{a:replicability}); models with smaller context exhibit minimal code simulation capabilities, and we thus excluded them from our analysis. Unless otherwise specified, we ran, for each benchmark, three independent runs with $30$ programs each.
Finally, we emphasise that we only evaluate models without access to compilers/interpreters. By doing so, the evaluation focuses on the capacity of each model to turn a prompt specified in natural language into a routine and then simulate it. 
We initially evaluate each model with standard Chain of Thought~\cite{wei2022chain} (CoT). The best open- and closed-source models are then tested with advanced techniques, including Self-consistency (SC)~\cite{wang2023selfconsistency} and our prompting method, namely CoSm. 
Furthermore, in the Appendix (Sections~\ref{a:illustrations} and~\ref{a:errors}), we detail common failure cases, including the negative impact of prompt illustrations~\cite{kojima2022large} (e.g., k-shot prompting).

\section{Code Simulation}
This section describes our results with straight-line programs, code with critical paths, and approximate and fault-tolerant instructions.
We then study nested loops and sorting algorithms. The key controlled variable for the input is the number of instructions, in line with recent works in the area~\cite{zhou2023algorithms}. 

\subsection{Straight-line Programs Simulation}
\begin{figure*}
\centering
\includegraphics[width=0.8\linewidth]{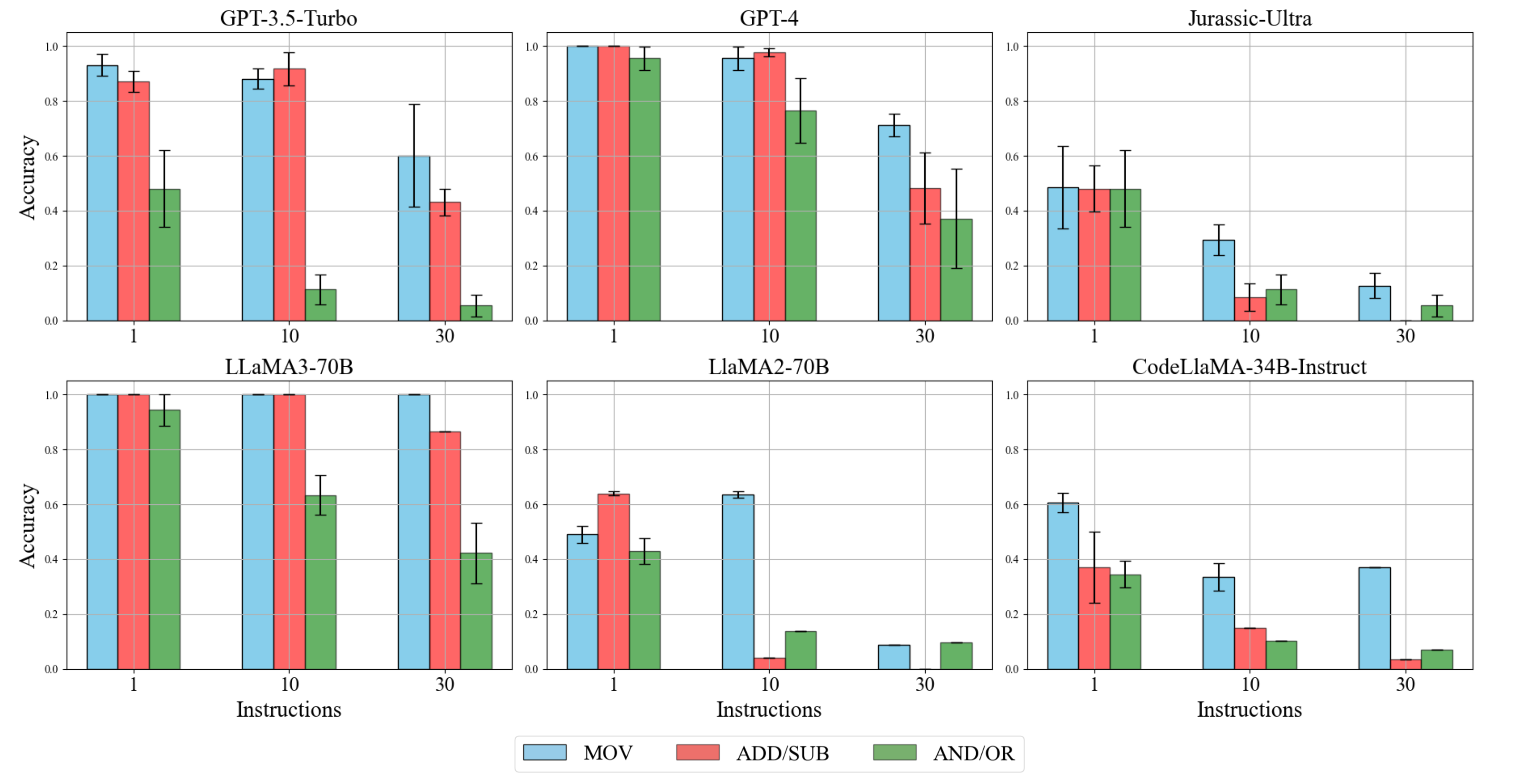}
\caption{Accuracy 
on $3$ independent runs of $30$ experiments each of different LLMs on code snippets with \textbf{solely} \{\texttt{and, or}\}, \{\texttt{add, sub}\} or \{\texttt{mov}\} instructions. We group results by codes of varying number of instructions (x-axis), namely $\{1, 10, 30\}$.}
\label{fig:llm-cpu-mixed-instructions}
\end{figure*}
\begin{figure*}
\centering
\includegraphics[width=\linewidth]{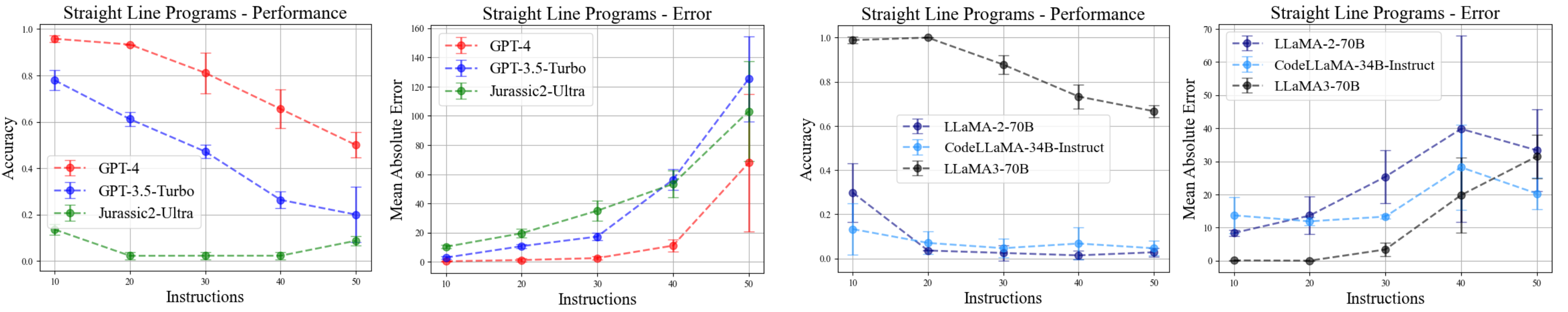}
\caption{Accuracy and Mean Absolute Error of different LLMs on code of varying length with \textbf{only} \{\texttt{add,sub}\} and \{\texttt{mov}\} instructions (out of $3$ independent runs of $30$ experiments each).}
\label{fig:llm-cpu-single-instructions}
\vspace{-4mm}
\end{figure*}
\begin{wrapfigure}{R}{0.3\textwidth}
\vspace{-5mm}
\begin{minipage}{0.3\textwidth}
\begin{lstlisting}
 a0=-1; a1=0; a2=-6
 a1 += a2
 a0 = a2
 a0 -= a0
 a1 = a0
 a0 -= a2
\end{lstlisting}
\vspace{-2mm}
\caption{Straight-line code.}\label{code:straight-line}
\end{minipage}
\vspace{-2mm}
\end{wrapfigure}
We first assess the simulation capabilities of different LLMs on code that contains only \{\texttt{add,sub}\}, \{\texttt{mov}\}, or logical-\{\texttt{and,or}\} instructions.
Figure~\ref{fig:llm-cpu-mixed-instructions} shows that for code containing only one type of instruction, Jurassic2-Ultra, LLaMA2-70B and CodeLLaMA-34b-Instruct are poor code simulators: their performance significantly downgrades with just $10$ instructions,
while GPT-3.5-Turbo, GPT-4 and LLaMA3-70B
are more accurate, though the same detrimental effect is evident, for example, on programs with $30$ sequential instructions. 
Logical instructions are hard to simulate for any model (green bar): we hypothesise that the reason is their low coverage in the training set since even a simple neural network can correctly compute logical-\{\texttt{and,or}\}, as we report in the Appendix~\ref{a:nn-and-or}.

Since the performance of any model considerably drops with logical-\{\texttt{and,or}\} instructions, we exclude such operation and synthesise straight-line programs with $\{10, 20, 30, 40, 50\}$ lines of instructions and a fixed number of variables (e.g., 5), as shown in Figure~\ref{code:straight-line}. We then prompt an LLM to compute the value of one of such variables at the end of the execution.
Both settings prompt each LLM to predict the state of a variable at the end of the computation.
Figure~\ref{fig:llm-cpu-single-instructions} shows our results for code with mixed instructions: in this task, they successfully achieve compositionality and reliably simulate code with mixed instructions.  GPT-4 and LLaMA3 are reliable instruction simulators, followed by GPT-3.5-Turbo. Conversely, Jurassic2-Ultra, LLaMA2-70B and CodeLLaMA-34B cannot simulate even short snippets of instructions.
Qualitatively, we further note that most errors occur when the output of the LLM consists only of the final result of the computation rather than the complete program execution trace.\footnote{We observed this phenomenon with GPT-4 in more than $95\%$ of cases, as per the code attachment.} Yet some notable exceptions occur, as described in the Appendix, Section~\ref{a:errors}.

\subsection{Smart Execution via Critical Paths}
\begin{wrapfigure}{R}{0.35\textwidth}
\vspace{-5mm}
\begin{minipage}{0.35\textwidth}
\begin{lstlisting}[escapechar=\%]
 a0 = a1 = a2 = 1
 %\textcolor{red}{a3 = a4 = a5 = -1}%
 a0 -= a1
 %\textcolor{red}{a3 -= a4}%
 %\textcolor{red}{a5 \&= a3}%
 %\textcolor{red}{a3 |= a5}%
 a0 += a1
 a1 -= a3
\end{lstlisting}
\vspace{-2mm}
\caption{Code with \textcolor{red}{critical path}. 
}\label{code:critical-path}
\end{minipage}
\vspace{-3mm}
\end{wrapfigure}
Some sequential problems can be solved without executing all the instructions in a program. For instance, consider the code in Figure~\ref{code:critical-path}, with a model prompted to predict the value of \verb~a3~.
To compute the value of \verb~a3~, it suffices to execute only those code blocks highlighted in \textcolor{red}{red}, which we refer to as the \textcolor{red}{critical path} of \verb~a3~.\footnote{From a theoretical perspective, a neural network that isolates a variable's critical path is straightforward to build, as shown in the Appendix, Section~\ref{a:critical-path}.}
We thus perform experiments with programs that contain critical paths shorter than the entire program.

\begin{figure*}
\centering
\includegraphics[width=\linewidth]{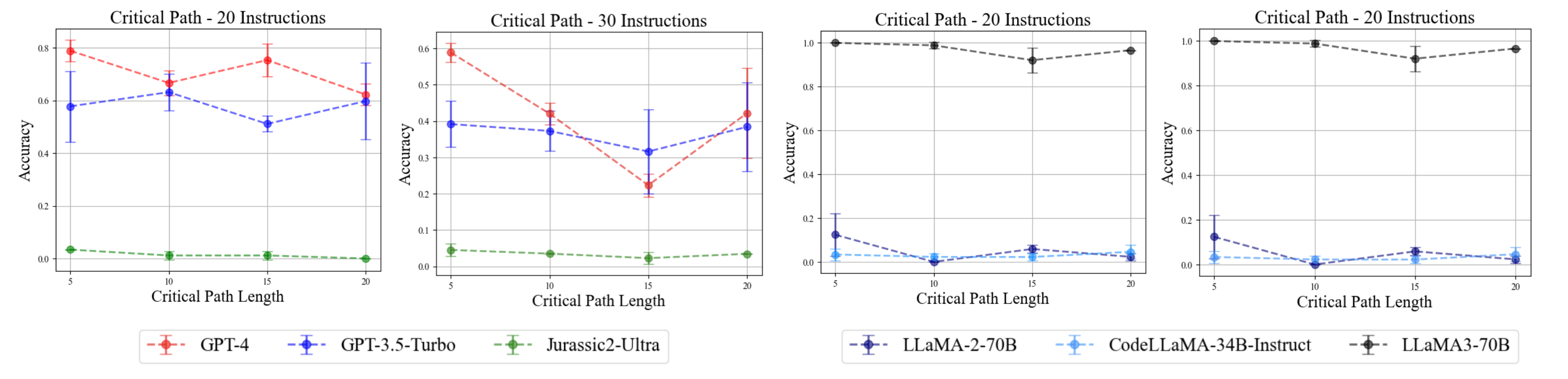}
\caption{Accuracy of different LLMs on $3$ runs of $30$ experiments each on programs with varying critical path lengths, for snippets of $20$ and $30$ lines of code respectively.}
\label{fig:code-complexity-critical-path}
\vspace{-5mm}
\end{figure*}

In Figure~\ref{fig:code-complexity-critical-path}, and for $3$ independent runs with $30$ programs each, we present the results when GPT-3.5-Turbo, GPT-4, Jurassic2-Ultra, LLaMA3-70B, LLaMA2-70B and CodeLLaMA-34b-Instruct are prompted to execute snippets of $20$ and $30$ instructions, with critical paths of varying length (i.e., $\{5, 10, 15, 20\}$).
GPT-4 and LLaMA3-70B can leverage smart execution, though LLaMA3-70B is better than GPT-4, especially on $30$ lines of code.
Although GPT-4's general simulation accuracy is higher than GPT-3.5-Turbo, it is less robust to variations of critical path length, i.e., GPT-4 suffers from a more severe accuracy drop compared to GPT-3.5-Turbo when critical path length approaches that of the entire program.
We also notice that LLaMA2-70B, CodeLLaMA-34B and Jurassic2-Ultra cannot generally execute instructions reliably.
As with straight-line execution, we notice that most errors occur when the output trace contains only the result, not the code simulation. 

\subsection{Approximate and Fault Tolerant Computation}
\begin{figure}
\centering
\includegraphics[width=1\linewidth]{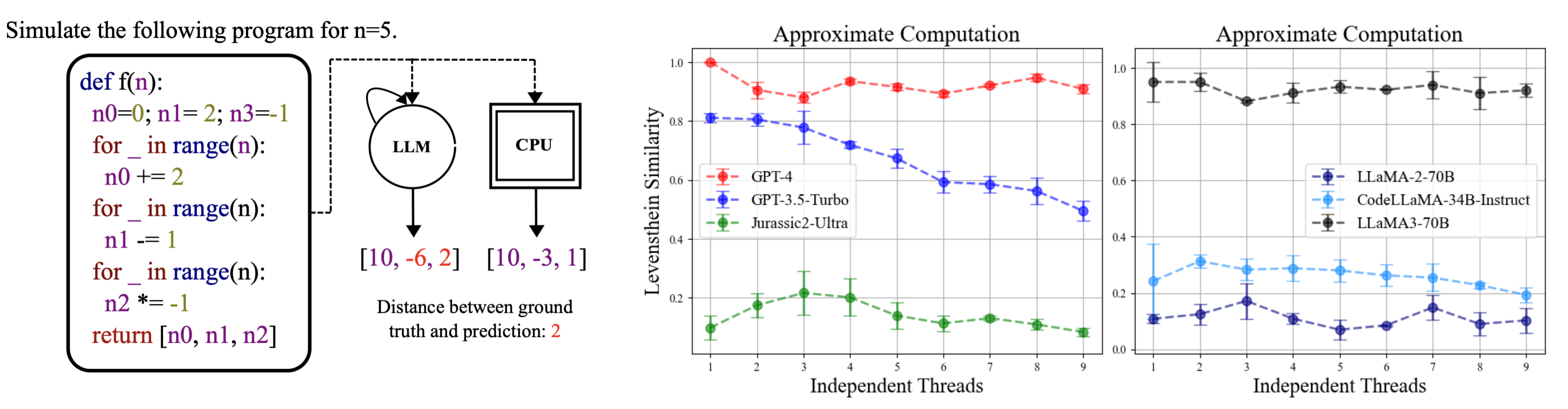}
\caption{On the left, an example of an algorithm to test the approximation capabilities of a model. On the right, the Levenshtein similarity between the ground truth and an LLM's output measures the performance of different models (the higher, the better).}
\label{fig:approximate-example}
\vspace{-5mm}
\end{figure} 
\textbf{Approximate computation} is evaluated with programs of $k$ \texttt{for} loops with $n$ instructions each. Each loop independently contributes to the final function return value, as shown in Figure~\ref{fig:approximate-example}. 
We denote by $\delta$ the probability of wrongly computing the result of each independent loop so that the probability of computing the exact result for a consistent \emph{analog} computer on a program is $(1-\delta)^k$
. For an LLM, $\delta$ is computed as the Levenshtein similarity between the ground truth values and the predicted results and is a proxy for the approximation capabilities on programs of varying complexity.
Results are reported in Figure~\ref{fig:approximate-example}.
GPT-4 and LLaMA3 are the best-performing models, with no accuracy degradation even for long programs with up to nine independent \emph{threads}.

\begin{figure}
\centering
\includegraphics[width=1\linewidth]{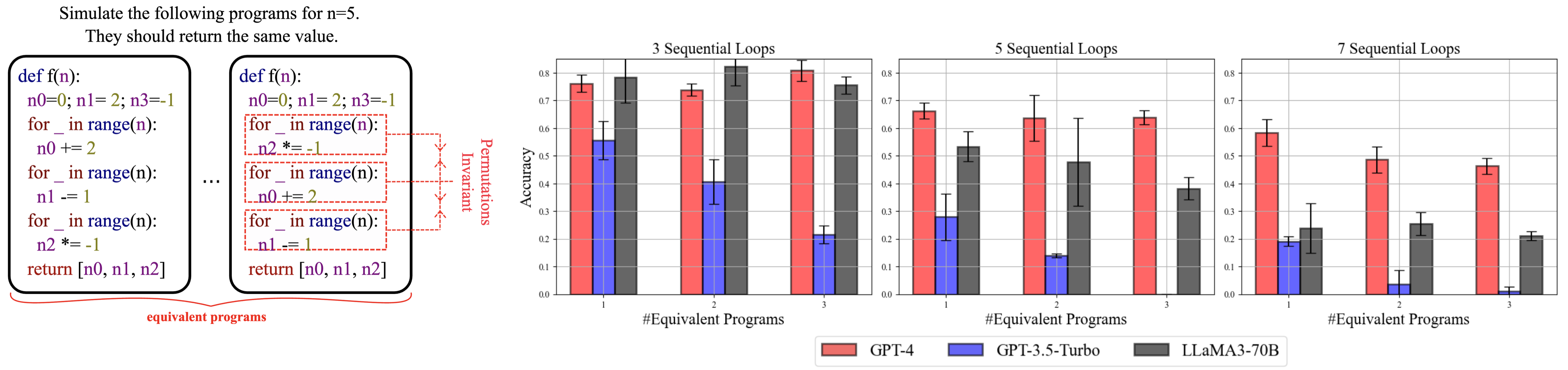}
\caption{Left: an example of a fault-tolerant algorithm. We feed an LLM with a few equivalent programs and instruct it to execute all of them to return the same result.
Right: how redundancy affects the performances of GPT-3.5-Turbo, GPT-4 and LLaMA3-70B on multiple equivalent programs.}
\label{fig:fault-tolerant}
\vspace{-5mm}
\end{figure}

A routine is \textbf{tolerant to faults} when it can recover from errors occurring during the computation. 
To test LLMs in this setting, we prompt a model with different variations of the same algorithm, specifying that the objective is to demonstrate they yield the same result.
Figure~\ref{fig:fault-tolerant} (left) reports an example of fault-tolerant prompts. The illustration is complemented by results for different LLMs on three independent runs of $30$ experiments each; the control variable is the number of equivalent programs fed to the model.
While redundancy neither alters GPT-4 accuracy nor improves its performance, GPT-3.5-Turbo is heavily affected by multiple equivalent programs in the prompt and experiences a severe decrease in performance. Results for Jurassic2-Ultra, LLaMA2-70B and CodeLLaMA-34B evidence low accuracy and are excluded from the evaluation (though inspectable in the code).



\subsection{Computational Complexity and Nested Loops}
\begin{wrapfigure}{R}{0.38\textwidth}
\begin{minipage}{0.38\textwidth}
\centering
\vspace{-2mm}
\includegraphics[width=1\linewidth]{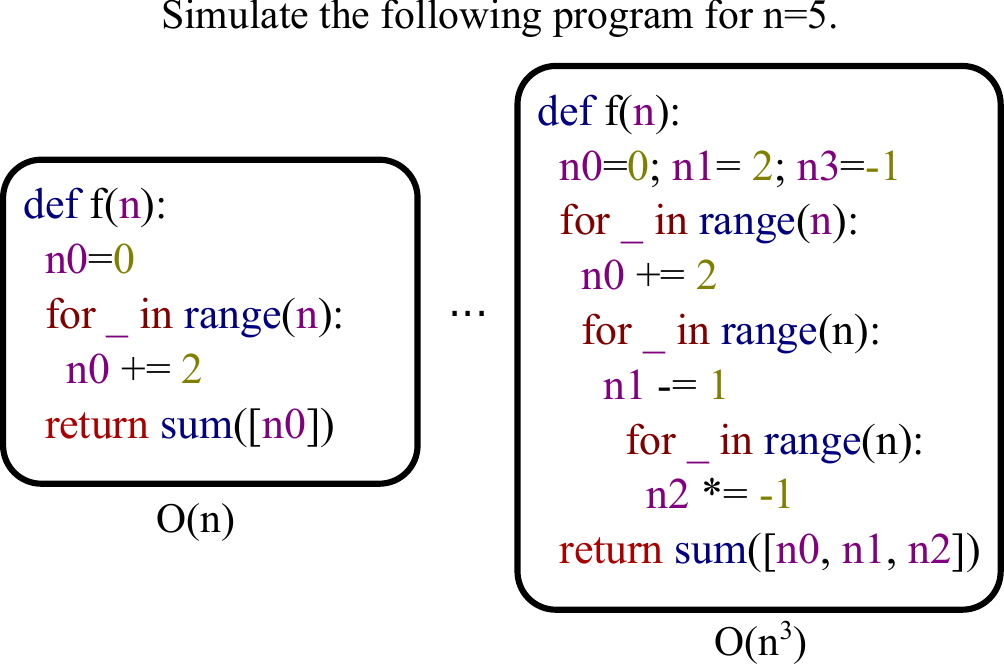}
\caption{Examples of programs with varying computational complexity: on the left, linear ($\mathcal{O}(n)$), on the right, cubic ($\mathcal{O}(n^3)$).}
\label{fig:nested-loops}
\end{minipage}
\vspace{-3mm}
\end{wrapfigure}
\begin{figure*}
\centering
\includegraphics[width=1\linewidth]{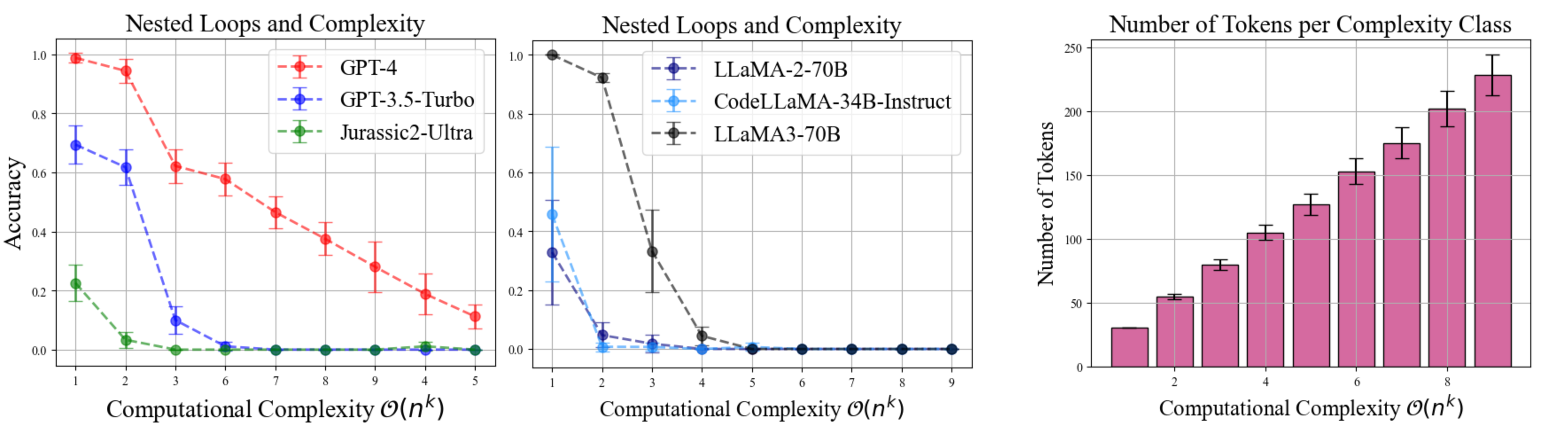}
\caption{Performances of different LLMs on nested loops with increasing computational complexity. On the right, the number of input tokens per complexity class grows linearly.}
\label{fig:complexity-error-res}
\vspace{-4mm}
\end{figure*}
\textbf{Nested loops} are a common instance of programs with polynomial running time $\mathcal{O}(n^k)$, where $k$ is the depth of loop nesting: see, e.g., Figure~\ref{fig:nested-loops}.
In this section, we prompt an LLM with programs that consist of $k$ nested loops with $n$ instructions each, i.e., with time complexity that ranges from $\mathcal{O}(n^{k=1})$ (linear) to $\mathcal{O}(n^{k=9})$. We measure the performance of a model to predict the exact result of the computation. 
By construction, the return value is an integer bounded between $\pm 2^k$, with overall upper- and lower-bounds bounded between $\pm 1024$, so we prevent high-order magnitude operands from influencing an LLM's performances. 
We run $3$ independent runs with batches of $30$ programs each and report the results for GPT-4, GPT-3.5-Turbo, Jurassic2-Ultra, LLaMA3-70B, LLaMA2-70B and CodeLLaMA-34b-Instruct.

Results in Figure~\ref{fig:complexity-error-res} evidence a \textbf{strong non-linear negative correlation} between the accuracy of GPT-3.5-Turbo, GPT-4 and LLaMA3-70B and the computational complexity of the function (right). In contrast, a strong linear correlation characterises the complexity of a function and its length (left). 
For high-performing LLMs (e.g., GPT-3.5, GPT-4, and LLaMA3-70B), algorithms whose complexity is beyond quadratic induce the most significant drop in performance. This suggests that the current state-of-the-art models cannot reliably simulate routines whose complexity is cubic or beyond. 
This phenomenon necessitates further investigations to connect the work in~\cite{zhou2023algorithms}, or other works on the computational capabilities of Transformers~\cite{weiss2021thinking}, with the computational complexity of a routine.  
Interestingly, LLaMA3-70B was the best-performing model on the straight-line, approximate and critical path code, yet on nested loops, GPT-4 outperforms it by a solid margin.
By inspecting the log results, we noticed that GPT-4, for high complexity programs (i.e., beyond $\mathcal{O}(n^2)$) implicitly unrolls the loops and \textbf{correctly guesses} the final result via \textbf{pattern matching}, surpassing any other model performance, including LLaMA3-70B. We report a comparison of LLaMA3-70B and GPT-4 in Appendix~\ref{a:gpt4-llama-loops}.

\subsection{Sorting Algorithms}\label{sec:sorting}
\begin{figure*}
    \centering
    \includegraphics[width=1\linewidth]{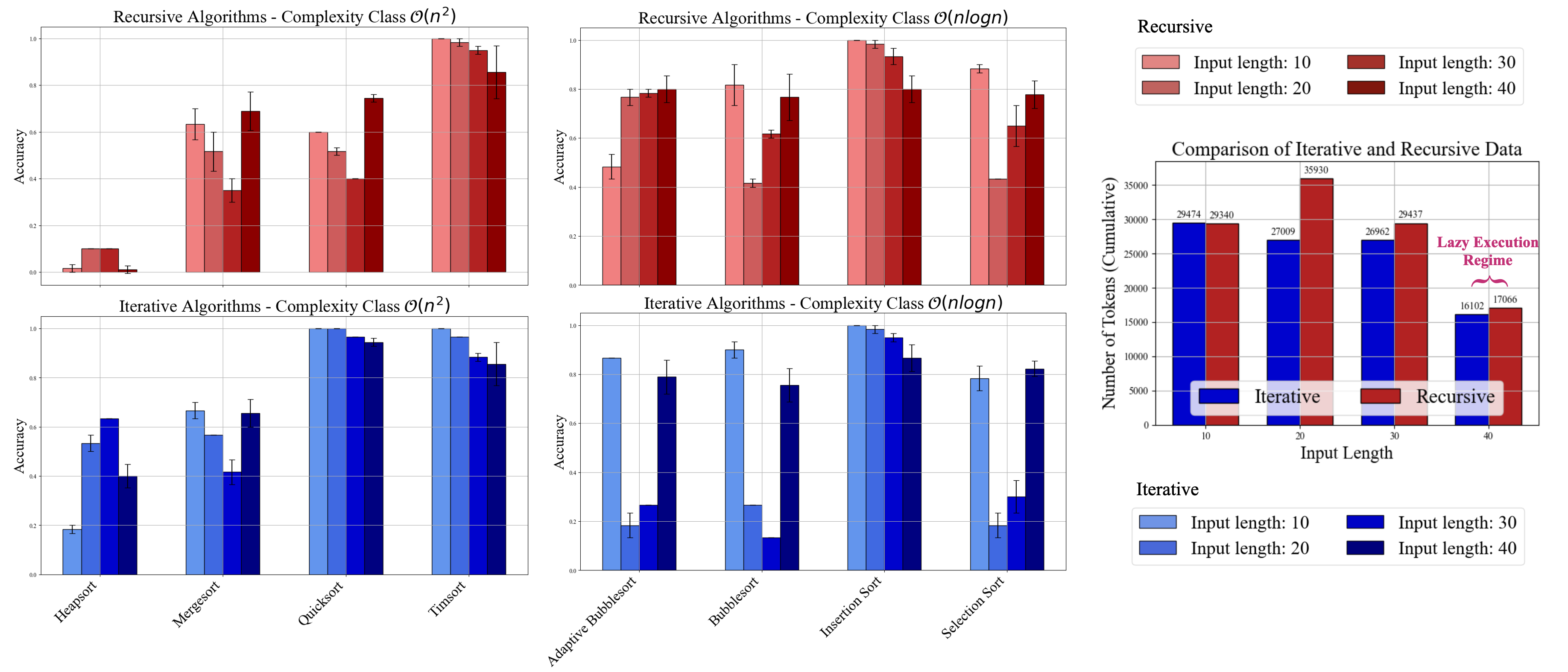}
    \caption{GPT-3.5-Turbo on sorting algorithms with varying complexity ($\mathcal{O}(n^2)$ and $\mathcal{O}(n \log n)$), both in their recursive (top) and iterative (bottom) versions. For long inputs, GPT-3.5 switches to a ``lazy execution regime'' (in \textbf{\textcolor{magenta}{magenta}}, right) where a model no longer simulates but just outputs the ordered sequence.}
    \label{fig:sort-iterative}
    \vspace{-3mm}
\end{figure*}
Next, we assessed whether LLMs can simulate \textbf{8 sorting routines} in their iterative and recursive versions and with log-linear or quadratic complexity (more details in the Appendix~\ref{a:sorting}). The control variable is the input length (i.e., to sort), which ranges between $\{10$, $20$, $30$, $40\}$ randomly sampled integers between $0$ and $100$. 

Results for GPT-3.5-Turbo and $3$ independent runs of $30$ experiments each are shown in Figure~\ref{fig:sort-iterative}. We analyse GPT-3.5-Turbo here as it results in the most interesting findings, though GPT-4 and LLaMA3-70B evaluations are reported in the Appendix,~\ref{a:sorting}.
Beyond confirming the implicit bias of Transformers towards ordered sequences~\cite{dufter2022position} (i.e., they tend to output ordered sequences), Figure~\ref{fig:sort-iterative} (left) shows that LLMs often provide the correct result for classic sorting algorithms such as Insertion Sort. However, with less common algorithms, LLMs trace the expected execution but fail even for vectors of few inputs.
In summary, we find that (i) standard sorting algorithms (e.g., Insertion and Quick Sort) lead to more accurate results; (ii) there is a weak correlation between algorithm complexity and the accuracy of its simulation, thus reinforcing the hypothesis that in this case, the model is not simulating the procedure; (iii) there is a weak correlation between the length of the input vector and the accuracy of a model. 

By contrast, GPT-3.5-Turbo is accurate on input vectors of max length (i.e., $40$). 
We name this phenomenon ``lazy execution regime'' (in \textbf{\textcolor{magenta}{magenta}}, Figure~\ref{fig:sort-iterative}, right): the number of tokens generated for long sequences is considerably smaller than for shorter inputs, hinting that the task's induction bias is more evident when the input is a consistent portion of the prompt. 
Finally, we document an endemic case of failure with LLMs such as GPT-3.5-Turbo and GPT-4. On standard algorithms such as Bubble Sort, long input sequences often lead to the wrong result when they contain repeated elements, as we document in Appendix~\ref{a:sort-failure}. We hypothesise that the probability of a sequence that contains repeated elements is so low, conditioned on what has been generated so far, that a model skips repeating elements, even when we set the `presence penalty' value to zero.\footnote{\url{https://platform.openai.com/docs/guides/text-generation/frequency-and-presence-penalties}}

\section{Eliciting Code Simulation}\label{sec:cosm}
\begin{wrapfigure}{R}{0.5\textwidth}
\vspace{-3mm}
\begin{minipage}{0.5\textwidth}
\begin{lstlisting}
"""
@code@
# 1. Simulate the above program instruction by instruction.
# 2. Report the trace of the program at the end of each iteration.
# 3. Think step by step and reply with the output of the function for the following input: @input@.
"""
\end{lstlisting}
\vspace{-3mm}
\caption{CoSm prompting technique. \texttt{@code@} and \texttt{@input@} tokens are replaced at prompt time.}\label{code:cosm}
\end{minipage}
\vspace{-2mm}
\end{wrapfigure}
\begin{figure}
\centering
\includegraphics[width=1\linewidth]{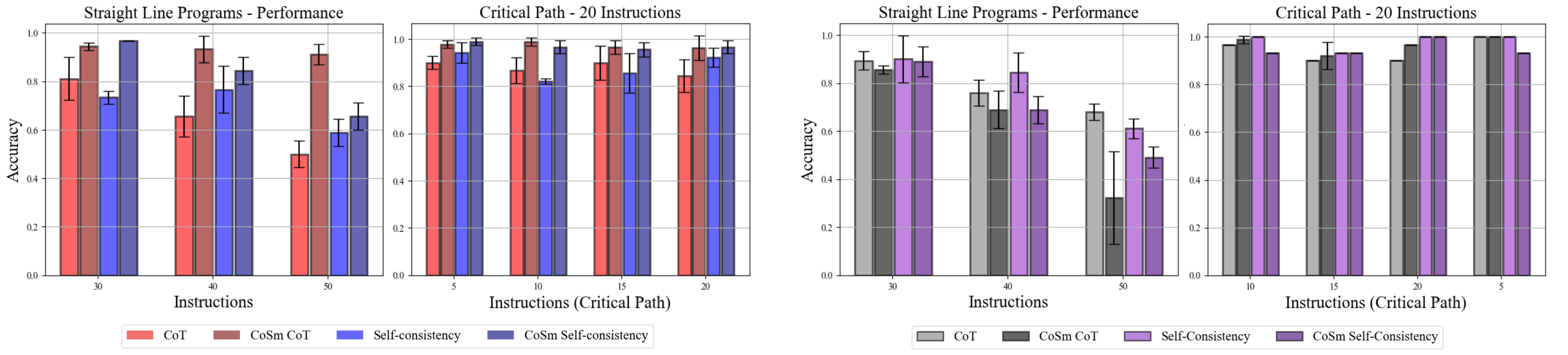}
\caption{CoSm improves the performance of GPT-4 in the straight-line and critical path (left) for both CoT and Self-consistency, while slightly decreases, or leaves unaltered, that of LLaMA3-70B (right).
}
\label{fig:cosm-straightline-critical-gpt4-llama}
\vspace{-3mm}
\end{figure}
\begin{figure}
\minipage{0.61\textwidth}
  \includegraphics[width=1.02\linewidth]{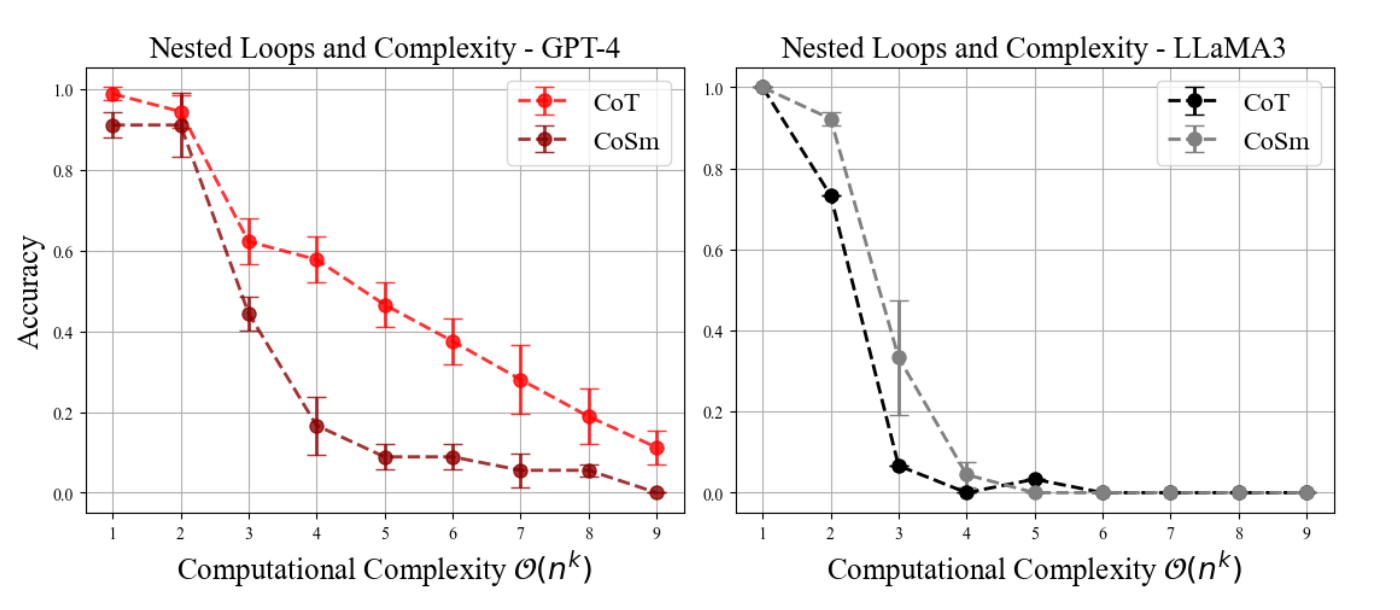}
\caption{On nested loops with varying computational complexity, CoSm guarantees improvements over CoT on LLaMA3-70B, while it decreases that of GPT-4, as it does not unroll the computation explicitly and uses pattern recognition to solve the task.}
\label{fig:cosm-complexity-gpt4}
\endminipage\hfill
\minipage{0.35\textwidth}
  \includegraphics[width=1.05\linewidth]{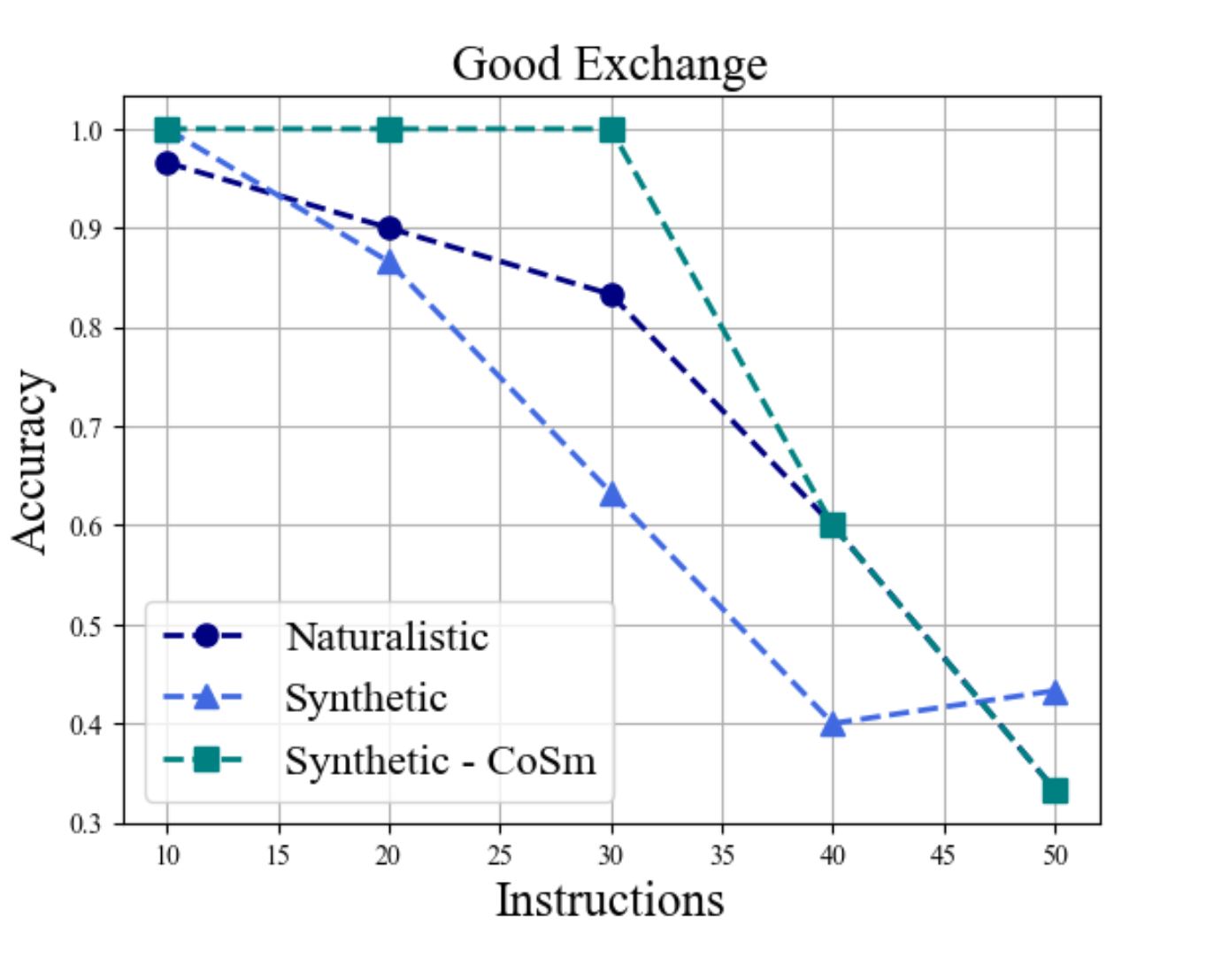}
\caption{GPT-4 on the good exchange task; CoSm in the \emph{synthetic} setting matches, for medium-length sequences, the performance of an oracle (e.g., a code interpreter).
}
\label{fig:cosm-good-exchange}
\endminipage
\vspace{-5mm}
\end{figure}

We propose Chain of Simulation (CoSm). As per the template in Figure~\ref{code:cosm}, this new, surprisingly simple yet effective prompting technique forces a model to simulate each instruction sequentially while keeping track of the program trace.
To prove CoSm's effectiveness, we select the most accurate models from the previous evaluation, GPT-4 and LLaMA3-70B, and assess whether CoSm further improves their performance when used with CoT and Self-Consistency.
For GPT-4, CoSm consistently improves code simulation capabilities in straight-line critical paths and nested loops, as per Figures~\ref{fig:cosm-straightline-critical-gpt4-llama} and~\ref{fig:cosm-complexity-gpt4}, while it degrades the performance on nested loops. On the other hand, the effect is the opposite on LLaMA3-70B, i.e., CoSm does not improve straight lines and critical paths but marginally improves nested loops. We believe that GPT-4 is better at leveraging patterns when prompted to perform simulation, e.g., it computes the result of nested loop operations without unrolling the computation.


We then revisit the good exchange task for GPT-4, where, surprisingly, the \emph{naturalistic} prompt elicits better performance than the \emph{synthetic} (see Figure~\ref{fig:cosm-good-exchange}). If we prompt the \emph{synthetic} setting with CoSm, GPT-4 outperforms the \emph{naturalistic} and the \emph{synthetic}, with oracle-like performance for medium-length sequences, showing that its simulation capabilities \textbf{can be elicited optimally} in this setting.  
\begin{figure}
\centering
    \includegraphics[width=1.\linewidth]{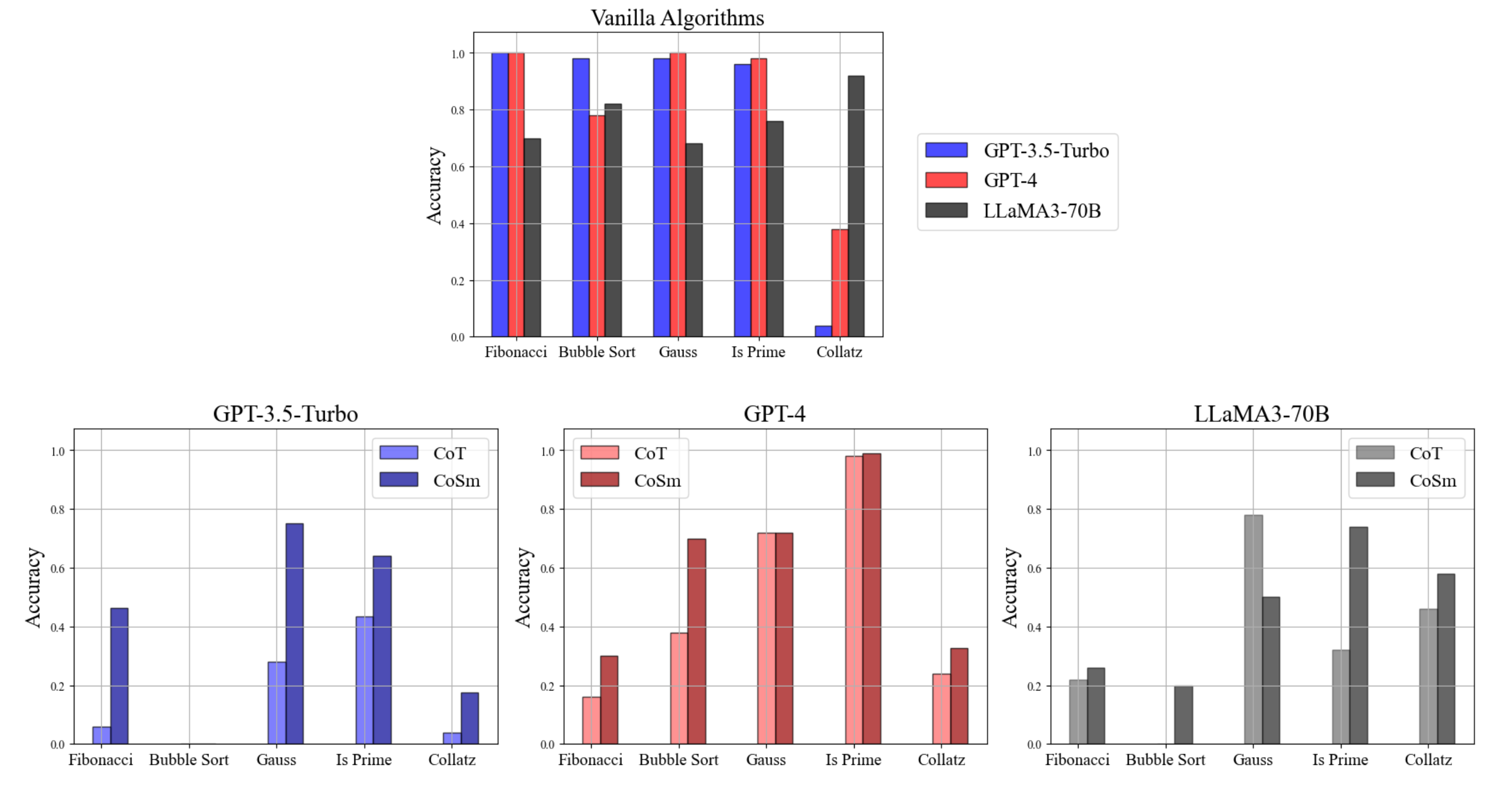}
    \caption{Results of GPT-3.5-Turbo, GPT-4 and LLaMA3-70B on $50$ independent simulations of classic algorithms and their variations. Top: the performance on the \emph{vanilla} implementation of each algorithm (with CoT). At the bottom, the performance on the variations for each model with standard CoT vs. CoSm (CoSm almost always improves them).}
    \label{fig:cosm}
\end{figure}
\begin{figure}
    \includegraphics[width=1.\linewidth]{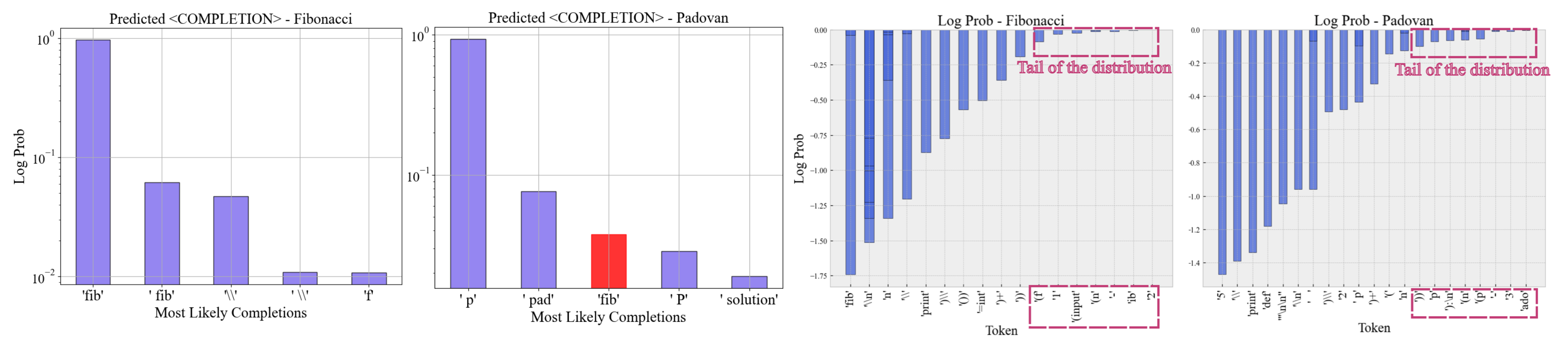}
    \caption{Auto-completion of GPT-3.5-Turbo-Instruct (left) when prompted to complete either the Fibonacci or Padovan function. On the right, a failure case of an anti-memorisation technique~\protect\cite{shi2023detecting}: the least likely predicted tokens of both Fibonacci and Padovan overlap and are not informative of memorisation.}
    \label{fig:barplot-fibo-pado}
    \vspace{-5mm}
\end{figure}
\newline
We concluded our study by investigating the tension between memorisation and code simulation and how CoSm beneficially prevents memorisation when detrimental to the downstream task.
We paired five standard algorithms with slight variations that neither affect the code length nor their computational complexity. Yet, their semantics are slightly changed to produce a different output.
We investigated the following algorithms and their variations: (i) the \textbf{Fibonacci} sequence paired with \textbf{Padovan}, a slight variation that modifies the return condition; (ii) \textbf{ascending Bubble Sort}, paired with the \textbf{descending} routine; (iii) the \textbf{Gauss algorithm}, paired with a variation that, instead of summing the first $n$ natural numbers, \textbf{adds even and subtracts odd numbers}; (iv) a \textbf{primality test} paired with the same routine on the \textbf{successor} of the input; (v) and the sum of the first $n$ \textbf{Collatz numbers}, paired with a variation that returns the sum of the \textbf{even numbers} in a Collatz sequence. 
All the functions and their variations were first anonymised to avoid bias towards known function names and implemented with the code that appears most frequently on GitHub (reported in the Appendix, Section~\ref{a:classic-algs}).
As shown in Figure~\ref{fig:cosm} (top), GPT-3.5-Turbo, GPT-4 and LLaMA3-70B are accurate on each classic algorithm, but their accuracy dropped significantly with the variations. On the other hand, their accuracy is marginally improved when prompted with CoSm (Figure~\ref{fig:cosm}, bottom).
We observe that LLMs anticipate the behaviour of a function by looking at specific dominating patterns. 
As shown in Figure~\ref{fig:barplot-fibo-pado} (right), GPT-3.5-Instruct completes a template of the Fibonacci function with tokens compatible with the function's scope. On Padovan, there is a non-negligible probability (the third most likely token, in \textbf{\textcolor{red}{red}}) that the predicted function is Fibonacci.
These results partially contradict the induction head mechanism~\cite{olsson2022context}, as a slight variation of a task where an LLM is accurate results in errors of large order of magnitude. Furthermore, this particular type of uncertainty goes unnoticed with methods that detect memorisation~\cite{shi2023detecting}, as shown in Figure~\ref{fig:barplot-fibo-pado} (right). While anti-memorisation techniques are effective with textual inputs (e.g., the Wikipedia dataset), they fail on code. We hypothesise that low-frequency tokens are informative for natural language but not on code as it is more structured, with low-likelihood tokens (in \textbf{\textcolor{magenta}{magenta}}) that are not predictive of a model's memorisation.

\section{Conclusions and Future Works}
We studied LLMs as simulators for digital devices, from straight-line code to high computational complexity algorithms. While the most powerful LLMs perform well on standard routines, they fail to handle more complex algorithms, highlighting their inherent limitations in being deployed as reliable digital simulators. 
To address these limitations, we introduced CoSm, a prompting method that improves LLM performance by minimising overreliance on shallow pattern recognition and memorization. CoSm bridges the gap between LLMs and practical applications in planning and digital computational models by providing a transferrable and interpretable technique and paves the way for LLMs to handle complex reasoning tasks more effectively.

\section*{Acknowledgments}
ELM is supported by the Alan Turing Institute.
AGC is supported by the Economic and Social Research Council (ESRC) under grant ES/W003473/1, by the Fundamental Research priority area of The Alan Turing Institute, and by the Turing Defence and Security programme through a partnership with the UK government in accordance with the framework agreement between GCHQ and the Alan Turing Institute. The support of Microsoft under their Accelerating Foundation Models Research initiative in providing Azure credits to AGC is gratefully acknowledged.

\clearpage

\bibliography{biblio}{}

\begin{thebibliography}{10}

\bibitem{llama3modelcard}
AI@Meta.
\newblock Llama 3 model card.
\newblock {\em technical report}, 2024.

\bibitem{berglund2023reversal}
Lukas Berglund, Meg Tong, Max Kaufmann, Mikita Balesni, Asa~Cooper Stickland, Tomasz Korbak, and Owain Evans.
\newblock The reversal curse: Llms trained on" a is b" fail to learn" b is a".
\newblock {\em arXiv preprint arXiv:2309.12288}, 2023.

\bibitem{biderman2023emergent}
Stella Biderman, USVSN~Sai Prashanth, Lintang Sutawika, Hailey Schoelkopf, Quentin Anthony, Shivanshu Purohit, and Edward Raf.
\newblock Emergent and predictable memorization in large language models.
\newblock {\em arXiv preprint arXiv:2304.11158}, 2023.

\bibitem{brown2020language}
Tom~B. Brown, Benjamin Mann, Nick Ryder, et~al.
\newblock Language models are few-shot learners.
\newblock In {\em Advances in Neural Information Processing Systems 33: Annual Conference on Neural Information Processing Systems 2020, NeurIPS 2020, December 6-12, 2020, virtual}, 2020.

\bibitem{chen2024evaluating}
Junkai Chen, Zhiyuan Pan, Xing Hu, Zhenhao Li, Ge~Li, and Xin Xia.
\newblock Evaluating large language models with runtime behavior of program execution, 2024.

\bibitem{chen2024language}
Minyu Chen, Guoqiang Li, Ling-I Wu, Ruibang Liu, Yuxin Su, Xi~Chang, and Jianxin Xue.
\newblock Can language models pretend solvers? logic code simulation with llms, 2024.

\bibitem{dufter2022position}
Philipp Dufter, Martin Schmitt, and Hinrich Sch{\"u}tze.
\newblock Position information in transformers: An overview.
\newblock {\em Computational Linguistics}, 48(3):733--763, 2022.

\bibitem{dziri2023faith}
Nouha Dziri, Ximing Lu, Melanie Sclar, Xiang~Lorraine Li, Liwei Jian, Bill~Yuchen Lin, Peter West, Chandra Bhagavatula, Ronan~Le Bras, Jena~D Hwang, et~al.
\newblock Faith and fate: Limits of transformers on compositionality.
\newblock {\em arXiv preprint arXiv:2305.18654}, 2023.

\bibitem{eldan2023s}
Ronen Eldan and Mark Russinovich.
\newblock Who’s harry potter? approximate unlearning in llms.
\newblock {\em arXiv preprint arXiv:2310.02238}, 2023.

\bibitem{frieder2023mathematical}
Simon Frieder, Luca Pinchetti, Alexis Chevalier, Ryan-Rhys Griffiths, Tommaso Salvatori, Thomas Lukasiewicz, Philipp~Christian Petersen, and Julius Berner.
\newblock Mathematical capabilities of {ChatGPT}.
\newblock {\em ArXiv preprint}, abs/2301.13867, 2023.

\bibitem{gao2020pile}
Leo Gao, Stella Biderman, Sid Black, Laurence Golding, Travis Hoppe, Charles Foster, Jason Phang, Horace He, Anish Thite, Noa Nabeshima, et~al.
\newblock The pile: An 800gb dataset of diverse text for language modeling.
\newblock {\em arXiv preprint arXiv:2101.00027}, 2020.

\bibitem{giannou2023looped}
Angeliki Giannou, Shashank Rajput, Jy-yong Sohn, Kangwook Lee, Jason~D Lee, and Dimitris Papailiopoulos.
\newblock Looped transformers as programmable computers.
\newblock {\em arXiv preprint arXiv:2301.13196}, 2023.

\bibitem{gu2024cruxeval}
Alex Gu, Baptiste Rozière, Hugh Leather, Armando Solar-Lezama, Gabriel Synnaeve, and Sida~I. Wang.
\newblock Cruxeval: A benchmark for code reasoning, understanding and execution, 2024.

\bibitem{hao2023reasoning}
Shibo Hao, Yi~Gu, Haodi Ma, Joshua~Jiahua Hong, Zhen Wang, Daisy~Zhe Wang, and Zhiting Hu.
\newblock Reasoning with language model is planning with world model.
\newblock {\em arXiv preprint arXiv:2305.14992}, 2023.

\bibitem{hou2023large}
Xinyi Hou, Yanjie Zhao, Yue Liu, Zhou Yang, Kailong Wang, Li~Li, Xiapu Luo, David Lo, John Grundy, and Haoyu Wang.
\newblock Large language models for software engineering: A systematic literature review.
\newblock {\em arXiv preprint arXiv:2308.10620}, 2023.

\bibitem{jojic2023gpt}
Ana Jojic, Zhen Wang, and Nebojsa Jojic.
\newblock Gpt is becoming a turing machine: Here are some ways to program it.
\newblock {\em arXiv preprint arXiv:2303.14310}, 2023.

\bibitem{kojima2022large}
Takeshi Kojima, Shixiang~Shane Gu, Machel Reid, Yutaka Matsuo, and Yusuke Iwasawa.
\newblock Large language models are zero-shot reasoners.
\newblock {\em ArXiv preprint}, abs/2205.11916, 2022.

\bibitem{la2023arrt}
Emanuele La~Malfa, Aleksandar Petrov, Simon Frieder, Christoph Weinhuber, Ryan Burnell, Anthony~G Cohn, Nigel Shadbolt, and Michael Wooldridge.
\newblock The arrt of language-models-as-a-service: Overview of a new paradigm and its challenges.
\newblock {\em arXiv preprint arXiv:2309.16573}, 2023.

\bibitem{liang2023code}
Jacky Liang, Wenlong Huang, Fei Xia, Peng Xu, Karol Hausman, Brian Ichter, Pete Florence, and Andy Zeng.
\newblock Code as policies: Language model programs for embodied control.
\newblock In {\em 2023 IEEE International Conference on Robotics and Automation (ICRA)}, pages 9493--9500. IEEE, 2023.

\bibitem{lieber2021jurassic}
Opher Lieber, Or~Sharir, Barak Lenz, and Yoav Shoham.
\newblock Jurassic-1: Technical details and evaluation.
\newblock {\em White Paper. AI21 Labs}, 1:9, 2021.

\bibitem{lin2024graphenhanced}
Fangru Lin, Emanuele~La Malfa, Valentin Hofmann, Elle~Michelle Yang, Anthony Cohn, and Janet~B. Pierrehumbert.
\newblock Graph-enhanced large language models in asynchronous plan reasoning, 2024.

\bibitem{liu2024codemind}
Changshu Liu, Shizhuo~Dylan Zhang, Ali~Reza Ibrahimzada, and Reyhaneh Jabbarvand.
\newblock Codemind: A framework to challenge large language models for code reasoning, 2024.

\bibitem{liu2023code}
Chenxiao Liu, Shuai Lu, Weizhu Chen, Daxin Jiang, Alexey Svyatkovskiy, Shengyu Fu, Neel Sundaresan, and Nan Duan.
\newblock Code execution with pre-trained language models.
\newblock {\em arXiv preprint arXiv:2305.05383}, 2023.

\bibitem{mccoy2023much}
R~Thomas McCoy, Paul Smolensky, Tal Linzen, Jianfeng Gao, and Asli Celikyilmaz.
\newblock How much do language models copy from their training data? {E}valuating linguistic novelty in text generation using {RAVEN}.
\newblock {\em Transactions of the Association for Computational Linguistics}, 11:652--670, 2023.

\bibitem{mccoy2023embers}
R~Thomas McCoy, Shunyu Yao, Dan Friedman, Matthew Hardy, and Thomas~L Griffiths.
\newblock Embers of autoregression: Understanding large language models through the problem they are trained to solve.
\newblock {\em arXiv preprint arXiv:2309.13638}, 2023.

\bibitem{min-etal-2022-rethinking}
Sewon Min, Xinxi Lyu, Ari Holtzman, et~al.
\newblock Rethinking the role of demonstrations: What makes in-context learning work?
\newblock In Yoav Goldberg, Zornitsa Kozareva, and Yue Zhang, editors, {\em Proceedings of the 2022 Conference on Empirical Methods in Natural Language Processing}, Abu Dhabi, United Arab Emirates, December 2022. Association for Computational Linguistics.

\bibitem{nye2021show}
Maxwell Nye, Anders~Johan Andreassen, Guy Gur-Ari, Henryk Michalewski, Jacob Austin, David Bieber, David Dohan, Aitor Lewkowycz, Maarten Bosma, David Luan, et~al.
\newblock Show your work: Scratchpads for intermediate computation with language models.
\newblock {\em arXiv preprint arXiv:2112.00114}, 2021.

\bibitem{olsson2022context}
Catherine Olsson, Nelson Elhage, Neel Nanda, et~al.
\newblock In-context learning and induction heads.
\newblock {\em Transformer Circuits Thread}, 2022.
\newblock https://transformer-circuits.pub/2022/in-context-learning-and-induction-heads/index.html.

\bibitem{OpenAI2023GPT4TR}
OpenAI.
\newblock {GPT-4} technical report.
\newblock {\em ArXiv preprint}, abs/2303.08774, 2023.

\bibitem{ouyang2022training}
Long Ouyang, Jeffrey Wu, Xu~Jiang, et~al.
\newblock Training language models to follow instructions with human feedback.
\newblock {\em Advances in Neural Information Processing Systems}, 35:27730--27744, 2022.

\bibitem{perez2021attention}
Jorge P{\'e}rez, Pablo Barcel{\'o}, and Javier Marinkovic.
\newblock Attention is turing complete.
\newblock {\em The Journal of Machine Learning Research}, 22(1):3463--3497, 2021.

\bibitem{petrov2023prompting}
Aleksandar Petrov, Philip~HS Torr, and Adel Bibi.
\newblock When do prompting and prefix-tuning work? a theory of capabilities and limitations.
\newblock {\em arXiv preprint arXiv:2310.19698}, 2023.

\bibitem{pmlr-v80-rabinowitz18a}
Neil Rabinowitz, Frank Perbet, Francis Song, Chiyuan Zhang, S.~M.~Ali Eslami, and Matthew Botvinick.
\newblock Machine theory of mind.
\newblock In Jennifer Dy and Andreas Krause, editors, {\em Proceedings of the 35th International Conference on Machine Learning}, volume~80 of {\em Proceedings of Machine Learning Research}, pages 4218--4227. PMLR, 10--15 Jul 2018.

\bibitem{roziere2023code}
Baptiste Rozi{\`e}re, Jonas Gehring, Fabian Gloeckle, Sten Sootla, Itai Gat, Xiaoqing~Ellen Tan, Yossi Adi, Jingyu Liu, Tal Remez, J{\'e}r{\'e}my Rapin, et~al.
\newblock Code {Llama}: Open foundation models for code.
\newblock {\em arXiv preprint arXiv:2308.12950}, 2023.

\bibitem{santos2023always}
Eddie~Antonio Santos, Prajish Prasad, and Brett~A Becker.
\newblock Always provide context: The effects of code context on programming error message enhancement.
\newblock In {\em Proceedings of the ACM Conference on Global Computing Education Vol 1}, pages 147--153, 2023.

\bibitem{scao2022bloom}
Teven~Le Scao, Angela Fan, Christopher Akiki, Ellie Pavlick, Suzana Ili{\'c}, Daniel Hesslow, Roman Castagn{\'e}, Alexandra~Sasha Luccioni, Fran{\c{c}}ois Yvon, Matthias Gall{\'e}, et~al.
\newblock {BLOOM}: A {176B}-parameter open-access multilingual language model.
\newblock {\em ArXiv preprint}, abs/2211.05100, 2022.

\bibitem{schuurmans2023memory}
Dale Schuurmans.
\newblock Memory augmented large language models are computationally universal.
\newblock {\em arXiv preprint arXiv:2301.04589}, 2023.

\bibitem{searle1980minds}
John~R Searle.
\newblock Minds, brains, and programs.
\newblock {\em Behavioral and brain sciences}, 3(3):417--424, 1980.

\bibitem{shanahan2022talking}
Murray Shanahan.
\newblock Talking about large language models.
\newblock {\em ArXiv preprint}, abs/2212.03551, 2022.

\bibitem{shi2023detecting}
Weijia Shi, Anirudh Ajith, Mengzhou Xia, Yangsibo Huang, Daogao Liu, Terra Blevins, Danqi Chen, and Luke Zettlemoyer.
\newblock Detecting pretraining data from large language models.
\newblock {\em arXiv preprint arXiv:2310.16789}, 2023.

\bibitem{touvron2023llama}
Hugo Touvron, Thibaut Lavril, Gautier Izacard, Xavier Martinet, Marie-Anne Lachaux, Timoth{\'e}e Lacroix, Baptiste Rozi{\`e}re, Naman Goyal, Eric Hambro, Faisal Azhar, et~al.
\newblock {LLaMA}: Open and efficient foundation language models.
\newblock {\em ArXiv preprint}, abs/2302.13971, 2023.

\bibitem{touvron2023llama2}
Hugo Touvron, Louis Martin, Kevin Stone, Peter Albert, Amjad Almahairi, Yasmine Babaei, Nikolay Bashlykov, Soumya Batra, Prajjwal Bhargava, Shruti Bhosale, et~al.
\newblock {LLaMA 2}: Open foundation and fine-tuned chat models.
\newblock {\em ArXiv preprint}, abs/2307.09288, 2023.

\bibitem{tufano2023predicting}
Michele Tufano, Shubham Chandel, Anisha Agarwal, Neel Sundaresan, and Colin Clement.
\newblock Predicting code coverage without execution.
\newblock {\em arXiv preprint arXiv:2307.13383}, 2023.

\bibitem{turpin2024language}
Miles Turpin, Julian Michael, Ethan Perez, and Samuel Bowman.
\newblock Language models don't always say what they think: unfaithful explanations in chain-of-thought prompting.
\newblock {\em Advances in Neural Information Processing Systems}, 36, 2024.

\bibitem{chen2021evaluating}
Priyan Vaithilingam, Tianyi Zhang, and Elena~L. Glassman.
\newblock Expectation vs. experience: Evaluating the usability of code generation tools powered by large language models.
\newblock In {\em {CHI} '22: {CHI} Conference on Human Factors in Computing Systems - 5 May 2022, Extended Abstracts}, pages 332:1--332:7. {ACM}, 2022.

\bibitem{wang2023selfconsistency}
Xuezhi Wang, Jason Wei, Dale Schuurmans, Quoc Le, Ed~Chi, Sharan Narang, Aakanksha Chowdhery, and Denny Zhou.
\newblock Self-consistency improves chain of thought reasoning in language models, 2023.

\bibitem{webb2023emergent}
Taylor Webb, Keith~J. Holyoak, and Hongjing Lu.
\newblock Emergent analogical reasoning in large language models, 2023.

\bibitem{wei2022statistically}
Colin Wei, Yining Chen, and Tengyu Ma.
\newblock Statistically meaningful approximation: a case study on approximating turing machines with transformers.
\newblock {\em Advances in Neural Information Processing Systems}, 35:12071--12083, 2022.

\bibitem{wei2022chain}
Jason Wei, Xuezhi Wang, Dale Schuurmans, Maarten Bosma, Ed~Chi, Quoc Le, and Denny Zhou.
\newblock Chain of thought prompting elicits reasoning in large language models.
\newblock {\em ArXiv preprint}, abs/2201.11903, 2022.

\bibitem{weiss2021thinking}
Gail Weiss, Yoav Goldberg, and Eran Yahav.
\newblock Thinking like transformers.
\newblock In {\em International Conference on Machine Learning}, pages 11080--11090. PMLR, 2021.

\bibitem{west2023generative}
Peter West, Ximing Lu, Nouha Dziri, Faeze Brahman, Linjie Li, Jena~D Hwang, Liwei Jiang, Jillian Fisher, Abhilasha Ravichander, Khyathi Chandu, et~al.
\newblock The generative ai paradox:" what it can create, it may not understand".
\newblock {\em arXiv preprint arXiv:2311.00059}, 2023.

\bibitem{widjojo2023addressing}
Patricia Widjojo and Christoph Treude.
\newblock Addressing compiler errors: Stack overflow or large language models?
\newblock {\em arXiv preprint arXiv:2307.10793}, 2023.

\bibitem{yang2023code}
Zhou Yang, Zhipeng Zhao, Chenyu Wang, Jieke Shi, Dongsun Kim, DongGyun Han, and David Lo.
\newblock What do code models memorize? an empirical study on large language models of code.
\newblock {\em arXiv preprint arXiv:2308.09932}, 2023.

\bibitem{yuan2023well}
Zheng Yuan, Hongyi Yuan, Chuanqi Tan, Wei Wang, and Songfang Huang.
\newblock How well do large language models perform in arithmetic tasks?
\newblock {\em arXiv preprint arXiv:2304.02015}, 2023.

\bibitem{zan2023large}
Daoguang Zan, Bei Chen, Fengji Zhang, Dianjie Lu, Bingchao Wu, Bei Guan, Wang Yongji, and Jian-Guang Lou.
\newblock Large language models meet nl2code: A survey.
\newblock In {\em Proceedings of the 61st Annual Meeting of the Association for Computational Linguistics (Volume 1: Long Papers)}, pages 7443--7464, 2023.

\bibitem{zhang2024transformerbased}
Dylan Zhang, Curt Tigges, Zory Zhang, Stella Biderman, Maxim Raginsky, and Talia Ringer.
\newblock Transformer-based models are not yet perfect at learning to emulate structural recursion, 2024.

\bibitem{zhang2023can}
Shizhuo~Dylan Zhang, Curt Tigges, Stella Biderman, Maxim Raginsky, and Talia Ringer.
\newblock Can transformers learn to solve problems recursively?
\newblock {\em arXiv preprint arXiv:2305.14699}, 2023.

\bibitem{zhou2023algorithms}
Hattie Zhou, Arwen Bradley, Etai Littwin, Noam Razin, Omid Saremi, Josh Susskind, Samy Bengio, and Preetum Nakkiran.
\newblock What algorithms can transformers learn? a study in length generalization.
\newblock {\em arXiv preprint arXiv:2310.16028}, 2023.

\end{thebibliography}
\bibliographystyle{plain}

\clearpage

\appendix

\section*{Appendix / supplemental material}
\section{Replicability}\label{a:replicability}
\paragraph{LLMs' parameters and replicability details.}
GPT-3.5 and GPT-4 were queried via the Microsoft Azure APIs (Model version: 0613), while Jurassic2-Ultra was queried via its API interface provided by AI21.\footnote{\url{https://docs.ai21.com/reference/python-sdk}} LLaMA3-70B was queried via GROQ APIs,\footnote{\url{https://console.groq.com/docs/models}} while LLaMA2-70B and CodeLLaMA-34B were queried via Huggingface Pro APIs.\footnote{\url{https://huggingface.co/blog/inference-pro}}

With LLMs like GPT-3.5-Turbo and GPT-4 that introduce concerns about the replicability of results~\cite{la2023arrt}, our experimental setup balanced performances with applicability.
We set a zero temperature for GPT-3.5-Turbo and GPT-4 and $0.1$ for LLaMA3-70B (as zero results in a division error).
Regarding LLaMA2-70B and CodeLLaMA-34B, we initially tested them with a temperature of $0.1$; then, we increased it to maximise performances to $0.7$ (with limited success).
All the implementation details (e.g., the maximum number of tokens generated, the \emph{top\_p}, and the functions to retrieve a model's response) are available in the code repository, file \texttt{`utils.py'}.

\paragraph{Logs of the experiments.}
All the logs of the experiments are available in the folder \texttt{`logs'}, grouped by model and benchmarking task.

\paragraph{Prompt templates.}
We implemented Chain of Thought as introduced in the original work by Wei et al.~\cite{wei2022chain}.
Self-consistency performs three independent queries to the same model, with temperature $0.1$, and then aggregates the final answer via majority vote. All the prompt templates are available in the folder \texttt{`prompts'}, grouped by benchmarking task.

\section{A Neural Network for Critical Paths}\label{a:critical-path} 
\begin{figure}
    \centering
    \includegraphics[width=0.8\linewidth]{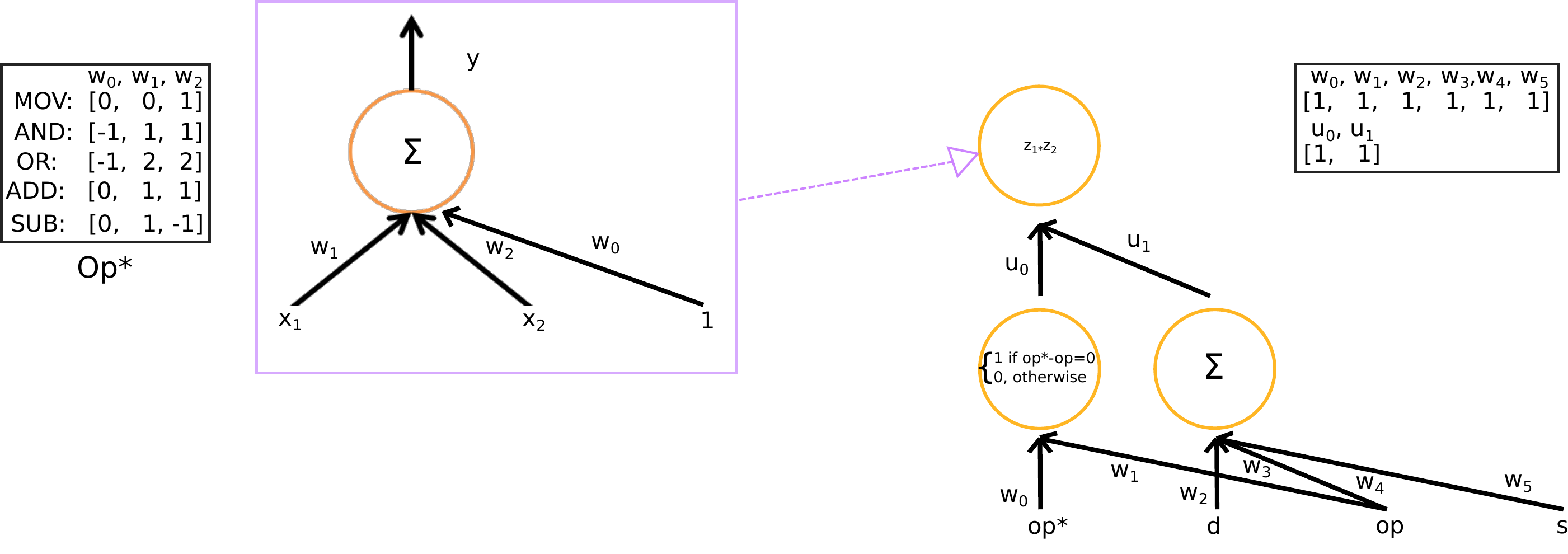}
    \caption{On the left, a linear neural network that computes \{\texttt{and, or}\}, \{\texttt{add, sub}\} and \{\texttt{mov}\} operations. On the right, a network executes a basic operation between two variables if and only if the operation is of a specific type (i.e., \{\texttt{mov}\}). For simplicity, the operation \texttt{op} is fed as an integer and compared to the ground truth value of \texttt{op}$^*$.}
    \label{fig:nn-basic-ops}
    \includegraphics[width=0.5\linewidth]{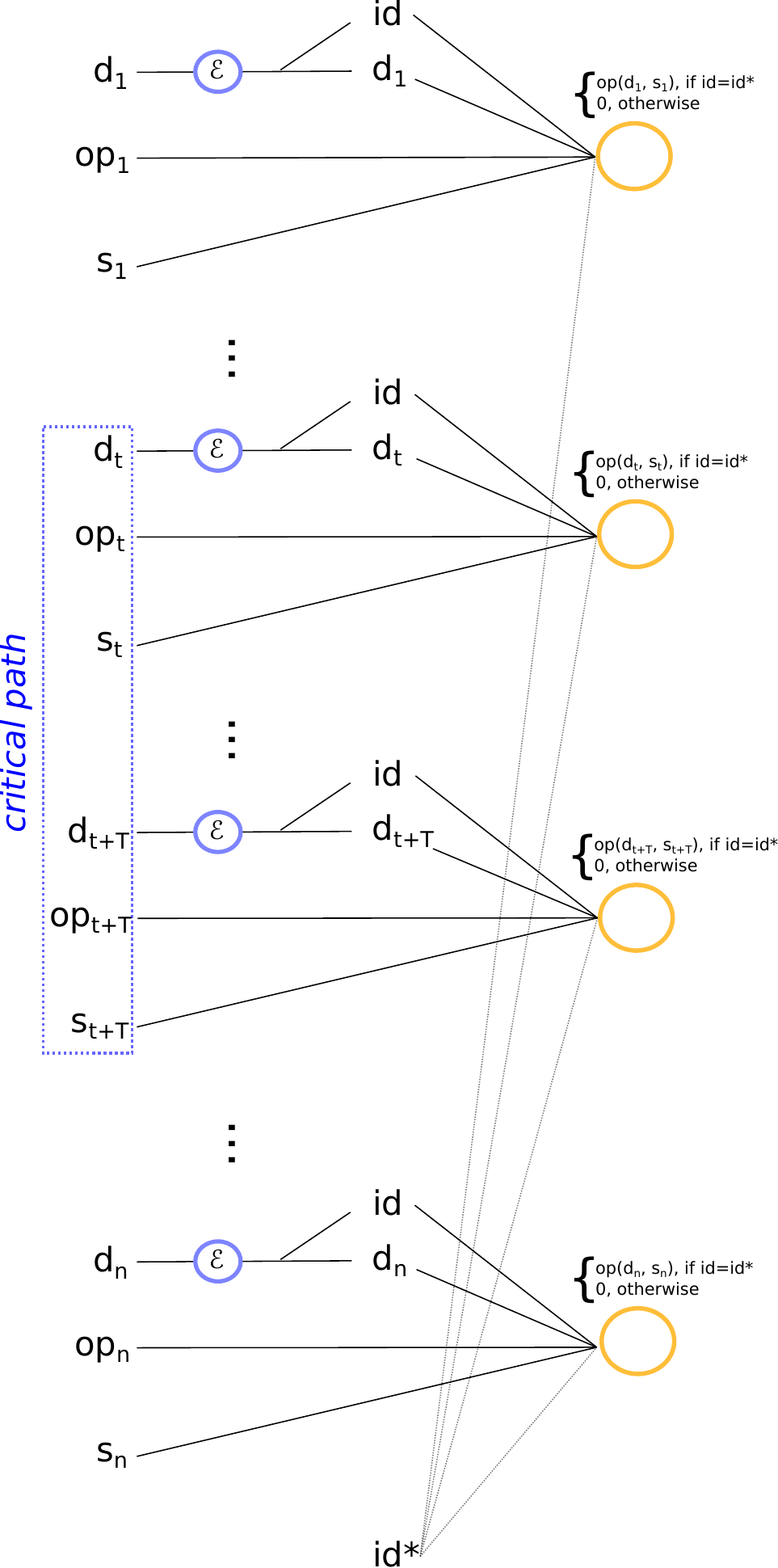}
    \caption{A network that identifies the \emph{critical path} of a variable. The embedding $\mathcal{E}$ duplicates each $d$estination variable into its numerical value and a unique identifier so that at the successive layer, the operation is executed if and only if there is a match between the target variable $id^*$ and the $d$estination.}
    \label{fig:nn-critical-path}
\end{figure}

Many articles have shown that neural networks can learn to precisely execute basic operations such as assignment, sum, etc.~\cite{giannou2023looped}. This means that, in principle, these models can simulate the internal operations of a CPU and thus behave like \emph{digital} computers.\footnote{That is true for finite length inputs, yet it is known neural networks and LLMs in general struggle with out-of-training samples~\cite{dziri2023faith}.} In Figure~\ref{fig:nn-basic-ops}, we sketch examples of feed-forward neural networks built to recognise and eventually execute the \{\texttt{and, or}\}, \{\texttt{add, sub}\} and \{\texttt{mov}\} operations between two variables.

\paragraph{Isolating the \emph{critical path} of a variable.}\label{a:nn-and-or}
We now provide constructive proof of a model that recognises a \emph{critical path} of any length and focuses only on that, ignoring the rest of the program. That model can be combined with the \{\texttt{and, or}\}, \{\texttt{add, sub}\} and \{\texttt{mov}\} ``heads" to execute operations in a CPU-fashion.
Suppose we have some code composed by $n>0$ operations in the form $(s, op, d)$, where $s$ and $d$ denote the $s$ource and $d$estination variables, while $op$ is either an \{\texttt{and, or}\}, \{\texttt{add, sub}\} or \{\texttt{mov}\} instruction. An example is the operation $a_0 += a_1$, where $s:=a_0$, $op:=+$, and $s:=a_1$. We want to identify the \emph{critical path} of the variable $a_i$. We build a feed-forward neural network that takes the program as input and, when it encounters the variable $a_i$, isolates all the successive operations until $a_i$ is no longer updated. This translates to a model whose weights implement an \texttt{if} statement that recognises $a_0$: that is achieved by first duplicating each operation such that each variable obtains a unique representation in terms of its value and a numerical identifier, say $(a_j, id_{a_j}), \ \nexists (a_j,a_k) \ s.t. \ id_{a_j}=id_{a_k}$. One can obtain that with an embedding and the identity transformation (to carry on the value of each variable). The next step involves recognising the \emph{critical path} of $a_i$: with each variable that has a unique numerical identifier, it is sufficient to set the condition that any operation is set to zero by the weights of the network if they do not involve the variable $a_j$ as $s$ource. The resulting output is the \emph{critical path} of $a_j$, which the model can leverage to execute only that portion of the program. Such a network is sketched in Figure~\ref{fig:nn-critical-path}.

\section{Additional Results}

\subsection{On the Contribution of Illustrations in Code Simulation}\label{a:illustrations}
\begin{figure}
\centering
\includegraphics[width=1\linewidth]{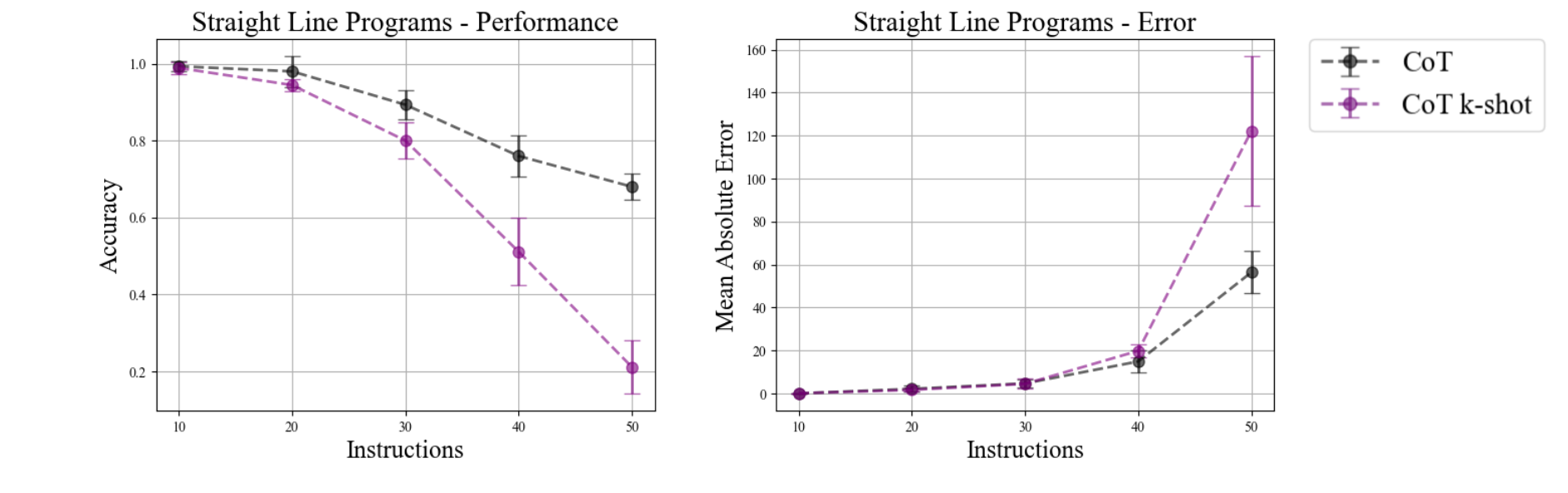}
\caption{Straight-line code simulation of LLaMA3-70B with CoT vs. CoT k-shot. K-shot degrades the model's performance, especially for long inputs, as the model runs out of context more rapidly.}
\label{fig:straight-line-kshot}
\end{figure} 
While k-shot prompting elicits superior performances on many tasks~\cite{brown2020language,touvron2023llama}, in our experience, and with two different approaches, it is ineffective with code simulation. 
The first approach includes, in the prompt, one to three illustration programs alongside their trace. The other approach explicitly instructs the model on how to simulate instructions (e.g., \{\texttt{add},\texttt{sub}\}).
Here's an example of an illustration that instructs a model on how to compute simple operations in a snippet of code:
\begin{lstlisting}
"""
Consider a Python program that can contain sums, assignments and logical operations.
A sum has the form a += b, which means that a becomes equal to its value plus that of b. A subtraction, in the form a -= b, works with the same rationale.
An assignment has the form a = b, which means that a becomes equal to b.
A logical and has the form a &= b, which means that a becomes equal to the logical and of a and b. A logical or, in the form a |= b, works with the same rationale.

Here's an example of a program that consists of 5 instructions, including assignments.
a0=5; a1=3; a2=8
a0 -= a1
a0 = a1
a4 += a1
a0 -= a2

The computation carried on is the following:
1. a0=5; a1=3; a2=8; a3=0; a4=4
2. a0 -= a1 -> a0 = 5 - 3 = 2
3. a0 = a1 -> a0 = 3
4. a4 += a1 -> a4 = 4 + 3 = 7
5. a0 -= a2 -> a0 = 3 - 8 = -5

So, for example, the value of a4 at the end of the iterations is 7 while the value of a0 is -5.
"""
\end{lstlisting}

While k-shot prompting is performed via a number ($1$, $2$, or $3$) of programs similar to the one below:
\begin{lstlisting}
"""
Consider an example of a program comprising 5 instructions, including assignments.
a0=5; a1=3; a2=8
a0 -= a1
a0 = a1
a4 += a1
a0 -= a2

The computation carried on is the following:
1. a0=5; a1=3; a2=8; a3=0; a4=4
2. a0 -= a1 -> a0 = 5 - 3 = 2
3. a0 = a1 -> a0 = 3
4. a4 += a1 -> a4 = 4 + 3 = 7
5. a0 -= a2 -> a0 = 3 - 8 = -5
"""
\end{lstlisting}

Both approaches are ineffective and do not elicit better performance, as Figure~\ref{fig:straight-line-kshot} and the code repository extensively reported (folder \texttt{`logs'}, models with \texttt{`kshot'} suffix.).
While this requires further experimental and theoretical validation, we explain this phenomenon based on the overwhelming evidence we gathered from our experiments and recent works in literature~\cite{min-etal-2022-rethinking,petrov2023prompting}. In particular, prompting examples are effective when they provide the model with (1)~the label space, (2)~the distribution of the input text, and (3)~the format of the sequence. In our evaluation, we discovered that $k$-shot is ineffective in our setting since (1) and (2) are explicit and already have high coverage in the training data. At the same time, the format sequence (3) consists of elementary operations for which a "template" does not help and can saturate a model's context. 

\subsection{GPT-4 and LLaMA3-70B on Loops Unrolling}\label{a:gpt4-llama-loops}
\begin{figure}
\centering
\includegraphics[width=1\linewidth]{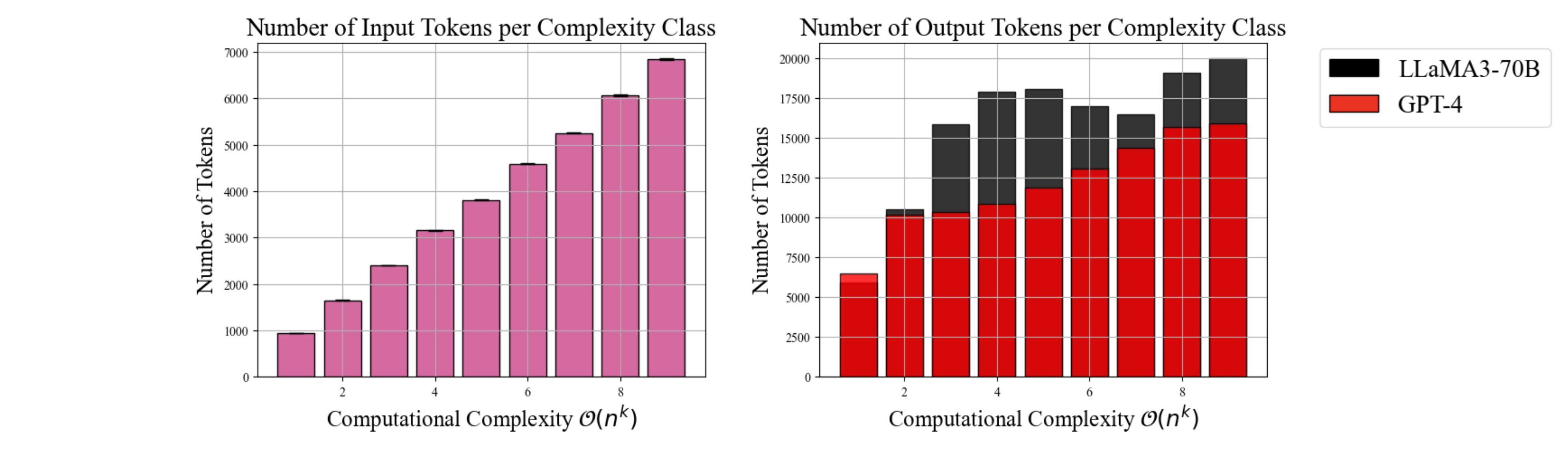}
\caption{On sorting algorithms, the number of input tokens grows linearly (left). At the same time, GPT-4 outputs fewer tokens than LLaMA3-70B, especially for complexity larger than $\mathcal{O}(n^2)$. The number is approximately the same for linear complexity. Both the graphs report the \textbf{cumulative} number of tokens over $30$ experiments.}
\label{fig:nested-loops-tokens}
\end{figure} 

To give empirical evidence that GPT-4 does implicit computation without unrolling the loops, while LLaMA3-70B tries to execute each instruction sequentially, we computed the number of tokens each model outputs in response to programs with different computational complexity.
As reported in Figure~\ref{fig:nested-loops-tokens}, the \textbf{cumulative} number of input tokens grows linearly (left). At the same time, GPT-4 outputs fewer tokens than LLaMA3-70B, especially for complexity larger than $\mathcal{O}(n^2)$. The number is approximately the same for linear and quadratic complexity. 

\subsection{Sorting Algorithms}\label{a:sorting}
\begin{figure*}
    \centering
    \includegraphics[width=1\linewidth]{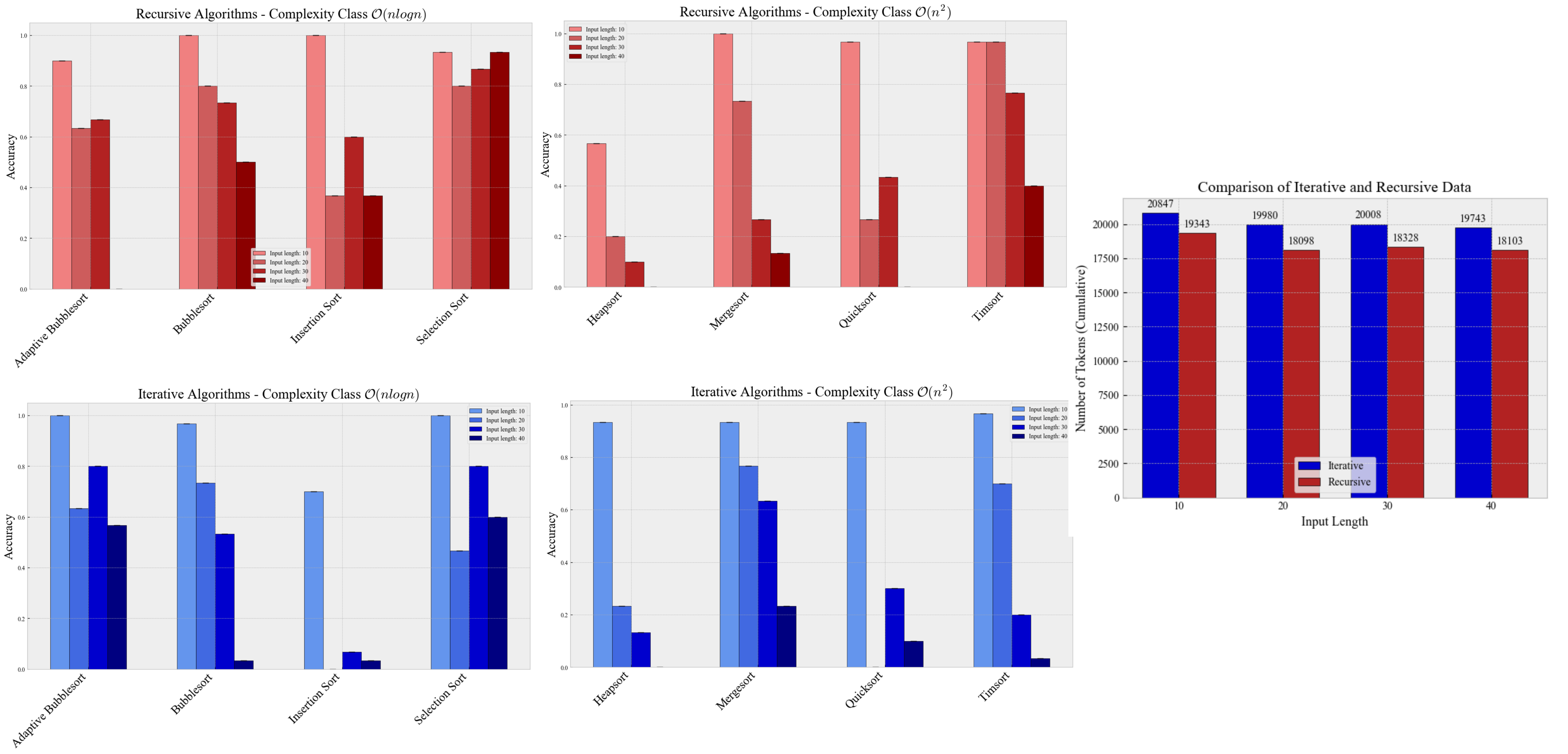}
    \caption{Results of LLaMA3-70B with CoT prompting technique on different sorting algorithms, both in their recursive (top) and iterative (bottom) versions. LLaMA3-70B is not a good simulator for sorting algorithms and, in general, does not understand the underlying task is sorting.}
    \label{fig:sort-iterative-llama-cot}
\end{figure*}

We report details on each sorting algorithm's space and time complexity in Table~\ref{tab:sorting-appendix}, while results for GPT-4 and LLaMA3-70B on all the sorting routines are reported in Figure~\ref{fig:sort-iterative-llama-cot} and~\ref{fig:sort-iterative-llama-cot}.
\begin{table*}
\centering
\caption{Sorting algorithms space and time complexity. They reference results in Figure~\ref{fig:sort-iterative}.}
\begin{adjustbox}{width=\textwidth} 
\begin{tabular}{|l|l|l|l|l|}
\hline
\textbf{Algorithm} & \textbf{Worst Time Complexity} & \textbf{Average Time Complexity} & \textbf{Best Time Complexity} & \textbf{Space Complexity} \\
\hline
\textcolor{black}{Insertion Sort} & $O(n^2)$ & $\Theta(n^2)$ & $\Omega(n)$ & It: $O(1)$ Rec: $O(n)$ \\
\hline
\textcolor{black}{Selection Sort}  & $O(n^2)$ & $\Theta(n^2)$ & $\Omega(n^2)$ & It: $O(1)$ Rec: $O(n)$ \\
\hline
\textcolor{black}{Bubblesort}  & $O(n^2)$ & $\Theta(n^2)$ & $\Omega(n^2)$ & It: $O(1)$ Rec: $O(n)$ \\
\hline
\textcolor{black}{Adaptive Bubblesort}  & $O(n^2)$ & $\Theta(n^2)$ & $\Omega(n)$ & It: $O(1)$ Rec: $O(n)$ \\
\hline
\textcolor{black}{Quicksort} & $O(n^2)$ & $\Theta(n \log(n))$ & $\Omega(n \log(n))$ & It: $O(n)$ Rec: $O(n)$ \\
\hline
\textcolor{black}{Mergesort}  & $O(n \log(n))$ & $\Theta(n \log(n))$ & $\Omega(n \log(n))$ & It: $O(n)$ Rec: $O(n)$ \\
\hline
\textcolor{black}{Timsort}  & $O(n \log(n))$ & $\Theta(n \log(n))$ & $\Omega(n)$ & It: $O(1)$ Rec: $O(n)$ \\
\hline
\textcolor{black}{Heapsort}  & $O(n \log(n))$ & $\Theta(n \log(n))$ & $\Omega(n \log(n))$ & It: $O(1)$ Rec: $O(\log n)$ \\
\hline
\end{tabular}\label{tab:sorting-appendix}
\end{adjustbox}
\end{table*}

\begin{figure*}
    \centering
    \includegraphics[width=1\linewidth]{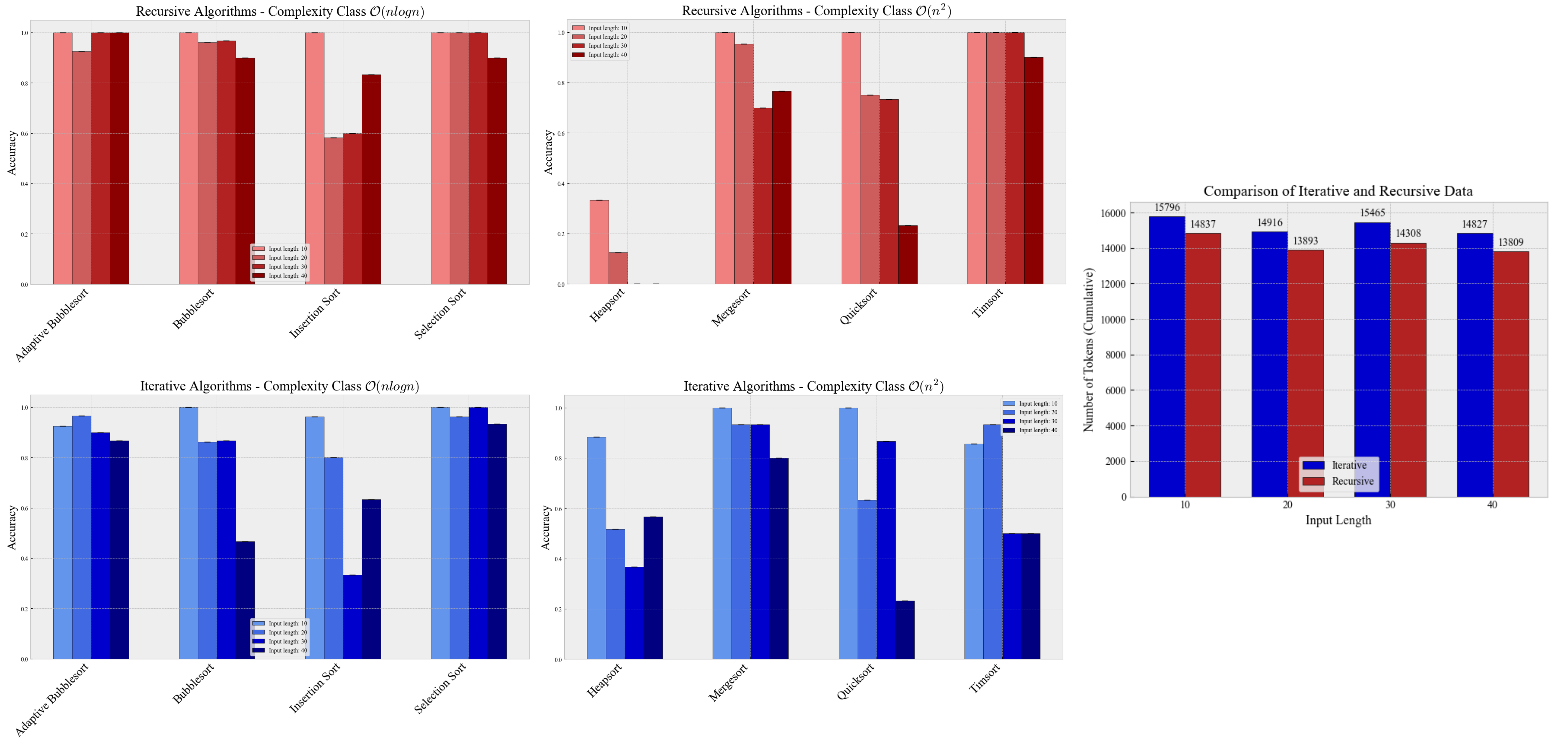}
    \caption{On top, results of GPT-4-Turbo with Chain of Thought prompting technique on different sorting algorithms, both in their recursive (top) and iterative (bottom) versions. Differently from GPT-3.5-Turbo, GPT-4 forces a model to simulate a routine and does not suffer from ``lazy execution" for longer input vectors (as illustrated in Figure~\ref{fig:sort-iterative} and the relative section).}
    \label{fig:sort-iterative-gpt-cot}
\end{figure*}

\subsubsection{A Case of Failure for Sorting}~\label{a:sort-failure}
We report a case of emblematic failure that appears frequently with LLMs such as GPT-3.5-Turbo and GPT-4. 
\begin{lstlisting}[escapechar=\%]
[85, %\textbf{\textcolor{magenta}{58}}%, 6, %\textbf{\textcolor{magenta}{58}}%, 34, %\textbf{\textcolor{magenta}{58}}%, 93, 47, 5, 89, 86, 12, 51, 76, 0, 3, 63, 6, 74, 52, 46, 61, 34, 92, 50, 56, 21, 25, %\textbf{\textcolor{magenta}{58}}%, 80].
\end{lstlisting}
The previous sequence, alongside the code for Bubble Sort, is fed to GPT-4. The input vector contains the number $58$ four times, highlighted in \textbf{\textcolor{magenta}{magenta}}). While GPT-3.5-Turbo correctly sorts the input vector, it reports the value $58$ thrice. The LLM output is reported below, with the sequence of interest highlighted in \textbf{\textcolor{magenta}{magenta}}:
\begin{lstlisting}[escapechar=\%]
[0, 3, 5, 6, 6, 12, 21, 25, 34, 34, 46, 47, 50, 51, 52, 56, %\textbf{\textcolor{magenta}{58}}%, %\textbf{\textcolor{magenta}{58}}%, %\textbf{\textcolor{magenta}{58}}%, 61, 63, 74, 76, 80, 85, 86, 89, 92, 93]
\end{lstlisting}
Our hypothesis is that the probability of a sequence that contains the number $58$ four times is so low, conditioned on what has been generated so far, that the model skips one of them, even though we set the presence penalty value to zero.
In this case, which is not isolated and appears frequently for similar inputs,  code simulation and correctness are in tension with the LLM output's probability distribution. 

\section{CoSm: Additional Results}

\subsection{CoSm Negative Results}\label{a:more-results-cosm}
An interesting failure case of LLMs is the execution of the descending Bubble Sort routine. 
Consider the \emph{vanilla} Bubble Sort routine, which sorts an array in ascending order:
\begin{lstlisting}[escapechar=\%]
def main(collection, size=0):
    length = len(collection)
    for i in reversed(range(length)):
        for j in range(i):
            if collection[j] %\textbf{\textcolor{red}{>}}% collection[j + 1]:
                collection[j], collection[j + 1] = collection[j + 1], collection[j]
    return collection
\end{lstlisting}
CoSm improves the Chain of Thought on Bubble Sort when the condition that swaps two elements in the input array is made explicit (as highlighted in \textbf{\textcolor{red}{red}}) by changing the operator \textbf{\textcolor{red}{>}} with \textbf{\textcolor{red}{<}}), as reported in the main paper, Figure~\ref{fig:cosm}. The code is reported below:
\begin{lstlisting}[escapechar=\%]
def main(collection, size=0):
    length = len(collection)
    for i in reversed(range(length)):
        for j in range(i):
            if collection[j] %\textbf{\textcolor{red}{<}}% collection[j + 1]:
                collection[j], collection[j + 1] = collection[j + 1], collection[j]
    return collection
\end{lstlisting}
On the other hand, when the the routine that swaps two elements is more convoluted, GPT-3.5-Turbo and GPT-4 have zero accuracy. We report the modified routine below:
\begin{lstlisting}[escapechar=\%]
def main(collection, size=0):
    length = len(collection)
    for i in reversed(range(length)):
        for j in range(i):
            %\textcolor{red}{collection[j] = max(collection[j], collection[j+1])}%
            %\textcolor{red}{collection[j+1] = min(collection[j], collection[j+1])}%
    return collection
\end{lstlisting}

\clearpage

\section{Notable Failure Cases of LLMs}\label{a:errors}
\subsection{Straight-line Programs Simulation}
Consider the following snippet of code:
\begin{lstlisting}
"""
Simulate this Python code.
"""
a0=-3; a1=1; a2=-7; a3=10; a4=-1
a4 = a4  # first instruction that modifies a4
a0 = a2
a3 += a3
a4 -= a2
a4 = a3  # last instruction that modifies a4
a0 += a2
a1 -= a3
a3 += a1
a3 += a1
a0 += a1
a0 -= a3
a3 = a3
a2 -= a0
a0 += a0
a1 -= a3
a2 = a2
a2 += a3
a2 -= a1
a3 += a1
a0 -= a2

"""
Think step by step and then reply with the numerical value a3 has at the end of the computation. Enclose the solution between <result></result> tags.
"""
\end{lstlisting}
When GPT-4 is prompted to compute the value of $a4$ at the end of the computation, it correctly computes the value of $20$ for $a3$, which is then assigned to $a4$ (highlighted in the code above), but then updates $a4$ based on the value $a3$ assumes at a later stage in the computation. We report the full output here:
\begin{lstlisting}
"""
I'm sorry for the confusion, but as an AI text-based model, I don't have the ability to execute Python code. However, I can help you understand what the code does.

The code you provided is a series of assignments and arithmetic operations on variables a0, a1, a2, a3, and a4. The value of a4 is last updated on the line "a4 = a3". After this line, a4 is not updated again, so its final value will be whatever the value of a3 is at that point in the code.

To find the value of a3 at that point, we need to follow the operations on a3:

- a3 starts with a value of 10.
- It is then doubled ("a3 += a3"), so it becomes 20.
- It is then increased by a1 ("a3 += a1"), so it becomes 21.
- It is increased by a1 again ("a3 += a1"), so it becomes 22.

So, at the point where "a4 = a3" is executed, a3 has a value of 22, and therefore a4 will also have a value of 22.

So, the final value of a4 is 22. 

Please note that this is a manual interpretation of the code and the actual result may vary if the code is run in a Python environment.
"""
\end{lstlisting}

\subsection{Approximate Simulation}
\noindent {Wrong Simulation (Implicit Code Simulation):}
\begin{lstlisting}
"""
Simulate this Python function.
"""
def f(n):
	n_0=-1; n_1=1; n_2=1; n_3=1; n_4=0
	for _ in range(n):
		n_0*=2
	for _ in range(n):
		n_1*=-2
	for _ in range(n):
		n_2*=-1
	for _ in range(n):
		n_3+=2
	for _ in range(n):
		n_4-=2
	return [n_0, n_1, n_2, n_3, n_4]
"""
Think step by step and then reply with the output of the function for n=10. Enclose the solution between <result></result> tags.
"""
\end{lstlisting}

\noindent {LLM output:}
\begin{lstlisting}
"""
<result>[-1024, 1024, -1, 21, -20]</result>
"""
\end{lstlisting}

\noindent {Correct Simulation (Explicit Code Simulation):}
\begin{lstlisting}
"""
Simulate this Python function.
"""
def f(n):
	n_0=-1; n_1=-1; n_2=1; n_3=0; n_4=1
	for _ in range(n):
		n_0+=2
	for _ in range(n):
		n_1+=2
	for _ in range(n):
		n_2+=1
	for _ in range(n):
		n_3+=1
	for _ in range(n):
		n_4*=-2
	return [n_0, n_1, n_2, n_3, n_4]
"""
Think step by step and then reply with the output of the function for n=10. Enclose the solution between <result></result> tags.
"""
\end{lstlisting}

\noindent {LLM Output:}
\begin{lstlisting}
"""
The function `f(n)` takes an input `n` and performs several operations on different variables. Let's go through the function step by step:

1. Initialize the variables `n_0`, `n_1`, `n_2`, `n_3`, and `n_4` with the values -1, -1, 1, 0, and 1 respectively.

2. Run a loop `n` times and increment `n_0` by 2 in each iteration.

3. Run another loop `n` times and increment `n_1` by 2 in each iteration.

4. Run another loop `n` times and increment `n_2` by 1 in each iteration.

5. Run another loop `n` times and increment `n_3` by 1 in each iteration.

6. Run another loop `n` times and multiply `n_4` by -2 in each iteration.

7. Return a list containing the values of `n_0`, `n_1`, `n_2`, `n_3`, and `n_4`.

Now, let's simulate this function for `n=10`:

1. Initialize `n_0`, `n_1`, `n_2`, `n_3`, and `n_4` with the values -1, -1, 1, 0, and 1 respectively.

2. Run the first loop 10 times, incrementing `n_0` by 2 in each iteration. After the loop, `n_0` will be 19.

3. Run the second loop 10 times, incrementing `n_1` by 2 in each iteration. After the loop, `n_1` will be 19.

4. Run the third loop 10 times, incrementing `n_2` by 1 in each iteration. After the loop, `n_2` will be 11.

5. Run the fourth loop 10 times, incrementing `n_3` by 1 in each iteration. After the loop, `n_3` will be 10.

6. Run the fifth loop 10 times, multiplying `n_4` by -2 in each iteration. After the loop, `n_4` will be -1024.

7. Return the list `[19, 19, 11, 10, -1024]`.

Therefore, the output of the function for `n=10` is:
<result>[19, 19, 11, 10, -1024]</result>
"""
\end{lstlisting}

\clearpage 

\section{Prompts and Programs}\label{a:prompts}
\subsection{Programs with Mov Instructions}\label{a:program-mov}
\begin{lstlisting}
"""Execute this Python code. Think step by step and then reply with the numerical value a3 has at the end of the computation.
"""

a0=-1; a1=0; a2=-6; a3=-5; a4=-3
a1 = a3
a3 = a3
a3 = a2
a1 = 3
a0 = a3
a3 = a0
a1 = a1
a2 = a0
a0 = a4
\end{lstlisting}

\subsection{Programs with Mixed Instructions}\label{a:program-mixed}
 
\begin{lstlisting}
"""Execute this Python code. Think step by step and then reply with the numerical value a2 has at the end of the computation.
"""

a0=-1; a1=0; a2=-6; a3=-5; a4=-3
a1 += a3
a3 = a3
a3 = a2
a1 &= a3
a4 += a0
a4 -= a1
a4 = a2
a4 += a4
a0 &= a3
a3 &= a0
a1 = a1
a2 &= a0
a0 &= a4
\end{lstlisting}

\subsection{Programs with a Critical Path}\label{a:program-critical-path}

\begin{lstlisting}[escapechar=\%]
"""Execute this Python code. Think step by step and then reply with the numerical value a9 has at the end of the computation.
"""

a0=-1; a1=-10; a2=-10; a3=1; a4=-6; 
%\textcolor{red}{a5=-10; a6=10; a7=9; a8=9; a9=10}%
a4 += a9
a3 = a3
a2 = a4
%\textcolor{red}{a9 \&= a7}% # begin critical path
%\textcolor{red}{a8 = a6}%
%\textcolor{red}{a9 \&= a5}%
%\textcolor{red}{a9 = a8}%  # end critical path
a5 += a9
a5 = a3
a2 = a9
\end{lstlisting}

\subsection{Code Complexity and Nested Loops}\label{a:nested-loops}
\begin{lstlisting}
"""Execute this Python function and report the numerical result for the input value n=10. Reply just with the solution.
"""

n0 = -1
def f(n):
    a = -2
    for _ in range(n):
      a += 2
    return a

\end{lstlisting}

\subsection{Sorting Algorithms}\label{a:sorting-code}
\subsubsection{Recursive Algorithms}
\noindent  Insertion Sort:
\begin{lstlisting}
def main(array, size, start=0):
    if start >= len(array) - 1:
        return array
    min_index = start
    for j in range(start + 1, len(array)):
        if array[j] < array[min_index]:
            min_index = j
    array[start], array[min_index] = array[min_index], array[start]
    return main(array, size, start + 1)
\end{lstlisting}

\noindent  Bubble Sort:
\begin{lstlisting}
def main(list_data, length) :
    for i in range(length - 1):
        if list_data[i] > list_data[i + 1]:
            list_data[i], list_data[i + 1] = list_data[i + 1], list_data[i]
    return list_data if length<2 else main(list_data, length - 1)
\end{lstlisting}

\noindent  Selection Sort:
\begin{lstlisting}
def main(array, size, start=0):
    if start >= len(array) - 1:
        return array
    min_index = start
    for j in range(start + 1, len(array)):
        if array[j] < array[min_index]:
            min_index = j
    array[start], array[min_index] = array[min_index], array[start]
    return main(array, size, start + 1)
\end{lstlisting}

\noindent  Adaptive Bubblesort:
\begin{lstlisting}
def main(list_data, length) :
    swapped = False
    for i in range(length - 1):
        if list_data[i] > list_data[i + 1]:
            list_data[i], list_data[i + 1] = list_data[i + 1], list_data[i]
            swapped = True
    return list_data if not swapped else main(list_data, length - 1)
\end{lstlisting}

\noindent  Quicksort:
\begin{lstlisting}
def main(array, high, low=0):
    if high==len(array):
        high=high-1
    if low < high:
        pi = f1(array, low, high)
        main(array,  pi - 1, low)
        main(array, high, pi + 1)
    return array
 
def f1(array, low, high):
    pivot = array[high]
    i = low - 1
    for j in range(low, high):
        if array[j] <= pivot:
            i = i + 1
            (array[i], array[j]) = (array[j], array[i])
    (array[i + 1], array[high]) = (array[high], array[i + 1])
    return i + 1
\end{lstlisting}

\noindent  Merge Sort:
\begin{lstlisting}
def main(arr, r, l=0):
    if r==len(arr):
        r=r-1
    if l < r:
        m = l+(r-l)//2
        main(arr, m, l)
        main(arr, r, m+1)
        f1(arr, l, m, r)
    return arr

def f1(arr, l, m, r):
    n1 = m - l + 1
    n2 = r - m
    L = [0] * (n1)
    R = [0] * (n2)
    for i in range(0, n1):
        L[i] = arr[l + i]
    for j in range(0, n2):
        R[j] = arr[m + 1 + j]
    i = 0
    j = 0
    k = l
    while i < n1 and j < n2:
        if L[i] <= R[j]:
            arr[k] = L[i]
            i += 1
        else:
            arr[k] = R[j]
            j += 1
        k += 1
    while i < n1:
        arr[k] = L[i]
        i += 1
        k += 1
    while j < n2:
        arr[k] = R[j]
        j += 1
        k += 1
\end{lstlisting}

\noindent  Tim Sort:
\begin{lstlisting}
def main(lst, size):
    length = len(lst)
    runs, s_runs = [], []
    new_run = [lst[0]]
    s_array = []
    i = 1
    while i < length:
        if lst[i] < lst[i - 1]:
            runs.append(new_run)
            new_run = [lst[i]]
        else:
            new_run.append(lst[i])
        i += 1
    runs.append(new_run)
    for run in runs:
        s_runs.append(f2(run))
    for run in s_runs:
        s_array = f1(s_array, run)
    return s_array

def f1(left, right):
    if not left:
        return right
    if not right:
        return left
    if left[0] < right[0]:
        return [left[0], *f1(left[1:], right)]
    return [right[0], *f1(left, right[1:])]

def f2(lst):
    length = len(lst)
    for index in range(1, length):
        value = lst[index]
        pos = f3(lst, value, 0, index - 1)
        lst = lst[:pos] + [value] + lst[pos:index] + lst[index + 1 :]
    return lst

def f3(lst, item, start, end):
    if start == end:
        return start if lst[start] > item else start + 1
    if start > end:
        return start
    mid = (start + end) // 2
    if lst[mid] < item:
        return f3(lst, item, mid + 1, end)
    elif lst[mid] > item:
        return f3(lst, item, start, mid - 1)
    else:
        return mid
\end{lstlisting}

\noindent  Heap Sort:
\begin{lstlisting}
def main(u_arr,size):
    n = len(u_arr)
    for i in range(n // 2 - 1, -1, -1):
        f1(u_arr, i, n)
    for i in range(n - 1, 0, -1):
        u_arr[0], u_arr[i] = u_arr[i], u_arr[0]
        f1(u_arr, 0, i)
    return u_arr

def f1(u_arr, index, heap_size):
    largest = index
    left_index = 2 * index + 1
    right_index = 2 * index + 2
    if left_index < heap_size and u_arr[left_index] > u_arr[largest]:
        largest = left_index

    if right_index < heap_size and u_arr[right_index] > u_arr[largest]:
        largest = right_index

    if largest != index:
        u_arr[largest], u_arr[index] = u_arr[index], u_arr[largest]
        f1(u_arr, largest, heap_size)
\end{lstlisting}

\subsubsection{Iterative Algorithms}
\noindent  Insertion Sort:
\begin{lstlisting}
def main(arr, size):
    for j, val in enumerate(arr[1:]):
        i = j
        while j >= 0 and val < arr[j]:
            arr[j + 1] = arr[j]
            j -= 1
        if j != i:
            arr[j + 1] = val
    return arr
\end{lstlisting}

\noindent  Bubble Sort:
\begin{lstlisting}
def main(collection, size=0):
    length = len(collection)
    for i in reversed(range(length)):
        for j in range(i):
            if collection[j] > collection[j + 1]:
                collection[j], collection[j + 1] = collection[j + 1], collection[j]
    return collection
\end{lstlisting}

\noindent  Selection Sort:
\begin{lstlisting}
def main(collection, size=0):
    length = len(collection)
    for i in reversed(range(length)):
        for j in range(i):
            if collection[j] > collection[j + 1]:
                collection[j], collection[j + 1] = collection[j + 1], collection[j]
    return collection
\end{lstlisting}

\noindent  Adaptive Bubblesort:
\begin{lstlisting}
def main(collection, size=0):
    length = len(collection)
    for i in reversed(range(length)):
        swapped = False
        for j in range(i):
            if collection[j] > collection[j + 1]:
                swapped = True
                collection[j], collection[j + 1] = collection[j + 1], collection[j]
        if not swapped:
            break
    return collection
\end{lstlisting}

\noindent  Quicksort:
\begin{lstlisting}
def main(arr, h, l=0):
    if h==len(arr):
        h=h-1
    size = h - l + 1
    stack = [0] * (size)
    top = -1
    top = top + 1
    stack[top] = l
    top = top + 1
    stack[top] = h
    while top >= 0:
        h = stack[top]
        top = top - 1
        l = stack[top]
        top = top - 1
        p = f1( arr, l, h )
        if p-1 > l:
            top = top + 1
            stack[top] = l
            top = top + 1
            stack[top] = p - 1
        if p + 1 < h:
            top = top + 1
            stack[top] = p + 1
            top = top + 1
            stack[top] = h
    return arr
\end{lstlisting}

\noindent  Merge Sort:
\begin{lstlisting}
def main(a, size):
    width = 1   
    n = len(a)                                         
    while (width < n):
        l=0;
        while (l < n):
            r = min(l+(width*2-1), n-1)        
            m = min(l+width-1,n-1)           
            f1(a, l, m, r)
            l += width*2
        width *= 2
    return a

def f1(a, l, m, r):
    n1 = m - l + 1
    n2 = r - m
    L = [0] * n1
    R = [0] * n2
    for i in range(0, n1):
        L[i] = a[l + i]
    for i in range(0, n2):
        R[i] = a[m + i + 1]
    i, j, k = 0, 0, l
    while i < n1 and j < n2:
        if L[i] <= R[j]:
            a[k] = L[i]
            i += 1
        else:
            a[k] = R[j]
            j += 1
        k += 1
    while i < n1:
        a[k] = L[i]
        i += 1
        k += 1
    while j < n2:
        a[k] = R[j]
        j += 1
        k += 1
\end{lstlisting}

\noindent  Tim Sort:
\begin{lstlisting}
def main(arr,n):
    min_run = 32
    n = len(arr)
    for i in range(0, n, min_run):
        f2(arr, i, min((i + min_run - 1), n - 1))
    size = min_run
    while size < n:
        for start in range(0, n, size * 2):
            middle = min((start + size - 1), (n - 1))
            end = min((start + size * 2 - 1), (n - 1))
            if middle < end:
                f1(arr, start, middle, end)
        size *= 2
    return arr
    
def f2(arr, left=0, right=None):
    if right is None:
        right = len(arr) - 1
    for i in range(left + 1, right + 1):
        key_item = arr[i]
        j = i - 1
        while j >= left and arr[j] > key_item:
            arr[j + 1] = arr[j]
            j -= 1
        arr[j + 1] = key_item

def f1(arr, left, middle, right):
    if arr[middle] <= arr[middle + 1]:
        return
    left_copy = arr[left:middle + 1]
    right_copy = arr[middle + 1:right + 1]
    left_copy_index = 0
    right_copy_index = 0
    s_index = left
    while left_copy_index < len(left_copy) and right_copy_index < len(right_copy):
        if left_copy[left_copy_index] <= right_copy[right_copy_index]:
            arr[s_index] = left_copy[left_copy_index]
            left_copy_index += 1
        else:
            arr[s_index] = right_copy[right_copy_index]
            right_copy_index += 1
        s_index += 1
    while left_copy_index < len(left_copy):
        arr[s_index] = left_copy[left_copy_index]
        left_copy_index += 1
        s_index += 1
    while right_copy_index < len(right_copy):
        arr[s_index] = right_copy[right_copy_index]
        right_copy_index += 1
        s_index += 1
\end{lstlisting}

\noindent  Heap Sort:
\begin{lstlisting}
def main(arr, n):
    f1(arr, n)
    for i in range(n - 1, 0, -1):
        arr[0], arr[i] = arr[i], arr[0]
        j, index = 0, 0
        while True:
            index = 2 * j + 1
            if (index < (i - 1) and
                arr[index] < arr[index + 1]):
                index += 1
            if index < i and arr[j] < arr[index]:
                arr[j], arr[index] = arr[index], arr[j]
            j = index
            if index >= i:
                break
    return arr

def f1(arr, n):
    for i in range(n):
        if arr[i] > arr[int((i - 1) / 2)]:
            j = i
            while arr[j] > arr[int((j - 1) / 2)]:
                (arr[j],
                 arr[int((j - 1) / 2)]) = (arr[int((j - 1) / 2)],arr[j])
                j = int((j - 1) / 2)
\end{lstlisting}

\subsection{Standard Algorithms and Variations}\label{a:classic-algs}
\noindent Fibonacci (iterative):
\begin{lstlisting}
def f(n):
    a, b = 0, 1
    if n <=1:
        return n       
    else:
        for i in range(1, n):
            c = a + b
            a = b
            b = c
        return b
\end{lstlisting}

\noindent Padovan (iterative):
\begin{lstlisting}
def g(n):
    a, b = 1, 1
    c, d = 1, 1
    for i in range(3, n+1):
        d = a + b
        a = b
        b = c
        c = d 
    return d
\end{lstlisting}

\noindent Bubble Sort (iterative):
\begin{lstlisting}
def f(v):
    n = len(v)
    for i in range(n):
        for j in range(0, n-i-1):
            if v[j] > v[j+1]:
                v[j], v[j+1] = v[j+1], v[j]
    return v
\end{lstlisting}

\noindent Bubble Sort Descending (iterative):
\begin{lstlisting}
def g(v):
    n = len(v)
    for i in range(n):
        for j in range(0, n-i-1):
            if 0 > v[j] - v[j+1]:
                v[j], v[j+1] = v[j+1], v[j]
    return v
\end{lstlisting}

\noindent Gauss Sum:
\begin{lstlisting}
def f(n):
    tot = 0
    for i in range(n):
        tot += i
    return tot
\end{lstlisting}

\noindent Gauss Sum and Subtraction:
\begin{lstlisting}
def g(n):
    tot = 0
    for i in range(n):
        tot += (i if i%2==0 else -i)
    return tot
\end{lstlisting}

\noindent Is Prime:
\begin{lstlisting}
def f(n):
    if n < 2: return False
    for x in range(2, int(n**0.5) + 1):
        if n % x == 0:
            return False
    return True
\end{lstlisting}

\noindent Is Prime on Successor:
\begin{lstlisting}
def g(n):
    n = n+1
    if n < 2: return False
    for x in range(2, int(n**0.5) + 1):
        if n % x == 0:
            return False
    return True
\end{lstlisting}

\noindent Collatz Sum Even:
\begin{lstlisting}
    def g(n):
        s = n
        while n != 1:
            if n % 2 == 0:
                n = n // 2
                s += n
            else:
                n = 3 * n + 1
        return s
\end{lstlisting}

\end{document}